\documentclass[Afour,sageh,times]{sagej}
\usepackage{amsmath}
\usepackage{amsmath, amssymb, amsthm}
\usepackage{algorithm}
\usepackage{algpseudocode}
\usepackage{geometry}
\usepackage{booktabs}
\usepackage{siunitx}
\usepackage{multirow}

\usepackage[table]{xcolor}

\usepackage{dsfont}
\usepackage{subfigure}
\usepackage{natbib}
\usepackage{array}
\usepackage{multirow}
\usepackage[utf8]{inputenc}
\usepackage[T1]{fontenc}
\usepackage{url}
\usepackage{booktabs}
\usepackage{amsfonts}
\usepackage{nicefrac}
\usepackage{microtype}
\usepackage{comment}
\usepackage{lipsum}
\usepackage{graphicx}
\usepackage{placeins}
\usepackage{bm}
\usepackage{bbm}
\usepackage{doi}
\newcommand{\xy}[1]{{\color{black}{#1}}}
\newcommand{\rev}[1]{{\color{black}{#1}}}
\newcommand{\pct}[1]{#1\! \%}

\newtheorem{proposition}{Proposition}
\theoremstyle{plain}

\theoremstyle{definition}

\newtheorem{assumption}{Assumption}

\definecolor{darkgreen}{RGB}{0,100,0}
\definecolor{tomato}{RGB}{255,99,71}
\definecolor{darkblue}{RGB}{0,0,139}
\usepackage{moreverb,url}

\newcommand\BibTeX{{\rmfamily B\kern-.05em \textsc{i\kern-.025em b}\kern-.08em
T\kern-.1667em\lower.7ex\hbox{E}\kern-.125emX}}

\def\volumeyear{2016}

\begin{document}

\runninghead{Carson and Chen}
\title{Precision autotuning for linear solvers via contextual bandit-based RL}
\author{\href{https://orcid.org/0000-0001-9469-7467}{\includegraphics[scale=0.06]{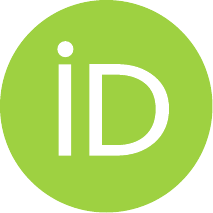}\hspace{1mm}Erin Carson}\affilnum{*} and \href{https://orcid.org/0000-0003-1778-393X}{\includegraphics[scale=0.06]{orcid.pdf}\hspace{1mm}Xinye Chen}\affilnum{$\dagger$}}
\affiliation{\affilnum{*}Charles University \\
Prague, Czech \\
\affilnum{$\dagger$}Sorbonne Université, CNRS, LIP6 \\
Paris, France}
\corrauth{Xinye Chen, Sorbonne Université, CNRS, LIP6, Paris, France}

\email{Xinye.Chen@lip6.fr}

\begin{abstract}
We propose a reinforcement learning (RL) framework for \xy{responsive} precision tuning for linear solvers, which can be extended to general algorithms. The framework is formulated as a contextual bandit problem and solved using incremental action-value estimation with a discretized state space to select optimal precision configurations for computational steps, \xy{retaining} precision and computational efficiency. To verify its effectiveness, we apply the framework to iterative refinement for solving linear systems $Ax = b$.  In this application, our approach dynamically chooses precisions based on calculated features from the system while maintaining acceptable accuracy and convergence. In detail, an action-value estimator takes discretized features (e.g., approximate condition number and matrix norm) as input and outputs estimated action values, from which a policy selects the actions (chosen precision configurations for specific steps), optimized via an $\epsilon$-greedy strategy to maximize a multi-objective reward to balance accuracy and computational cost. Empirical results demonstrate effective precision selection, \xy{increasing the use of lower-precision arithmetic} while maintaining accuracy comparable to double-precision baselines. \xy{We further evaluate the learned policies in a compiled CPU GMRES-IR implementation using FP16, FP32, and FP64 arithmetic for solver-level native validation.} The framework generalizes to diverse out-of-sample data and provides insights into applying RL precision selection to other numerical algorithms, advancing mixed-precision numerical methods in scientific computing. To the best of our knowledge, this is the first work on precision autotuning with RL with verification on unseen datasets.
\end{abstract}

\keywords{Mixed-precision computing, Reinforcement learning, Precision tuning, Linear solvers, AI for numerical methods}

\maketitle

\section{Introduction}
\label{sec:introduction}

Floating-point arithmetic \citep{8766229} is widely used in machine learning and scientific computing. With the rapid development of large language models \citep{DBLP:journals/corr/abs-2005-14165, hu2022lora, touvron2023} as well as large-scale GPU computing \citep{DBLP:journals/corr/abs-1710-03740, peng2023fp8lmtrainingfp8large}, numerical calculations and algorithm deployment with extensive use of floating-point arithmetic are often accompanied by a substantial increase in energy use. Therefore, using reduced-precision arithmetic in algorithms is typically required in many intensive computing kernels due to its lower communication, faster arithmetic, and energy efficiency. Seven commonly-used floating-point formats are referenced in \tablename~\ref{table:unitroundoff}. Modern accelerators provide substantially higher throughput for reduced precision than for FP64 arithmetic; for example, the NVIDIA H100 (NVIDIA's Hopper architecture) achieves up to 624 TFLOPS with FP8, offering an approximate $64\times$ peak-throughput ratio versus FP64 \citep{nvidia_fp8}. The work \cite{nvidia_fp8} demonstrates that the FP8 formats E4M3 and E5M2 can in some cases match FP16 accuracy with reduced compute and memory costs.

However, reduced-precision arithmetic often brings increased rounding errors and numerical instability, which can compromise the results of computations, especially at large scale \citep{10.1177/1094342016652462}.  The idea behind mixed-precision algorithms is to design an algorithm that combines low and high precision arithmetic while retaining an acceptable level of accuracy, which remains a major obstacle in modern approximate computing paradigms \citep{benkhalifa:hal-03978176, 7348659}. Accordingly, many theoretical analyses of rounding error, mixed-precision algorithms \citep{doi:10.1137/17M1122918, doi:10.1137/17M1140819, doi:10.1137/20M1316822, doi:10.1137/17M1122918, abhl24}, and precision tuning tools \citep{6877460,  10.1145/3213846.3213862} have been developed, which aim to achieve a balance between precision loss and arithmetic speedup.

Theoretical analyses of the use of multiprecision arithmetic and mixed-precision algorithms have been extensively conducted in the context of iterative refinement \citep{doi:10.1137/20M1316822, abhl24}, least squares problems \citep{doi:10.1137/19M1298263, doi:10.1137/20M1316822}, and GMRES \citep{jang:hal-05163845}. For example, iterative refinement can be executed in three precisions, but the convergence is only theoretically guaranteed for a limited range of problems, which can be overly restrictive in practice \citep{oktay2022multistage}. As such, one can instead turn to precision tuning tools to find a suitable mixed-precision approach. Existing tools typically require running the program multiple times via trial-and-error, and \xy{their scalability and generalization ability} are not verified, so they are either limited to small or medium-scale programs or suffer from efficiency issues.  An ideal case for a precision tuning tool or method is to fit the model in a limited number of training sets and learn the key features that can be used to infer mixed-precision configurations for new unseen data. These challenges need to be addressed to reduce the dependency on programmer efforts.

\begin{table*}[ht!]
\caption{Key parameters of seven floating-point formats; $u$ denotes the unit-roundoff corresponding to the precision, $x_{\min}$ denotes the smallest positive normalized floating-point number, $x_{\max}$ denotes the largest floating-point number, $t$ denotes the number of binary digits in the significand (including the implicit leading bit), $e_{\min}$ denotes exponent of $x_{\min}$, and $e_{\max}$ denotes exponent of $x_{\max}$.} \small
\label{table:unitroundoff}
\centering \setlength\tabcolsep{5pt}
\begin{tabular}{l l l l r r r}
\hline\\[-2.5mm]
& \qquad $u$      & \quad $x_{\min}$  &    \quad $x_{\max}$  &  $t$    &  $e_{\min}$  & $e_{\max}$\\[1mm] \hline\\[-0.6mm]
bfloat16 (BF16) & $3.91 \times 10^{-3}$ & $1.18 \times 10^{-38}$ & $3.39 \times 10^{38}$ & 8 & -126 & 127 \\
half precision (FP16) & $4.88 \times 10^{-4}$ & $6.10 \times 10^{-5}$ & $6.55 \times 10^{4}$ & 11 & -14 & 15 \\
TensorFloat-32 (TF32) & $9.77 \times 10^{-4}$ & $1.18 \times 10^{-38}$ & $1.70 \times 10^{38}$ & 11 & -126 & 127 \\
single (FP32) & $5.96 \times 10^{-8}$ & $1.18 \times 10^{-38}$ & $3.40 \times 10^{38}$ & 24 & -126 & 127 \\
double (FP64) & $1.11 \times 10^{-16}$ & $2.23 \times 10^{-308}$ & $1.80 \times 10^{308}$ & 53 & -1022 & 1023 \\[1mm]
\hline 
\end{tabular}
\end{table*}

A contextual bandit (also known as a multi-armed bandit with context) is a simplified reinforcement learning (RL) problem where the agent makes decisions based on the current context (i.e., state) without considering the future contexts or rewards of its actions; in other words, the agent optimizes for one-step (immediate) reward rather than long-term cumulative reward. In this paper, we propose a general framework for applying RL for precision selection in general numerical algorithms, formulated as a contextual bandit problem due to the independence of problem instances and the single decision per instance. This is the first work to formally define the use of RL for precision selection in any algorithms. The framework uses an epsilon-greedy strategy, controlled by a decaying epsilon determined by training episodes, to select precision levels for the prespecified algorithm steps, treating precision choices as a multi-armed bandit problem within a RL environment. The epsilon parameter balances exploration and exploitation, starting high to encourage diverse precision selections and decaying over time to favor optimal choices. We believe this framework enables at least three advantages:
\begin{enumerate}
    \item Identifying the precision level required for user-specific computational steps in terms of hand-crafted features or low computational complexity features modeled from linear systems;
    \item Finding salient features of the linear systems (e.g., condition number, matrix norm, non-zero values, diagonal dominance) to determine the reduced-precision configurations;
    \item Inferring reduced- and mixed-precision configurations for unseen datasets.
\end{enumerate}

We apply this framework to the GMRES-based iterative refinement (GMRES-IR) method \cite{doi:10.1137/17M1122918} to verify the effectiveness and generalization ability of RL for precision selection, aiming to achieve a trade-off between the use of low precision and accuracy. The framework is particularly suited for scenarios with limited training data and is designed to generalize to new data, and can be easily implemented in an online learning routine to avoid model retraining.

The rest of the paper is structured as follows. Section~\ref{sec:related_work} discusses the existing literature on precision autotuning as well as its pros and cons, and a comparison with our work. Section~\ref{sec:general_framework} presents our general framework for precision selection;
to verify its effectiveness, we use the GMRES-IR method, a well-studied mixed-precision method, as a case study detailed in Section~\ref{sec:gmres_application}. Section~\ref{sec:exps} presents the experiments and empirical results on dense, randomly sparse, and PDE-discretized matrices under varying condition numbers and sparsity. Section~\ref{sec:conclusion} discusses limitations as well as future work.

\section{Related Work}\label{sec:related_work}
There exist a variety of theoretical rounding error analysis of basic mixed-precision algorithms aimed at understanding the underlying arithmetic behavior;  see the survey \cite{doi:10.1177/10943420211003313} and references therein. However, for real-world problems, which can be more complex, the scientific community typically uses more general approaches such as precision tuning tools, which operate at program compile time and runtime, immediately resulting in reduced-precision programs produced automatically.  In this section, we briefly review existing approaches for the analysis of mixed-precision algorithms and automated precision tuning that are closely related to our method.

Theoretical analysis aims to design reduced-precision settings for specific steps and analyze their rounding error so that the reduced-precision arithmetic can be used safely under certain assumptions to retain algorithm accuracy and numerical stability. Analyses in \cite{WILKINSON1966} and \cite{10.1145/321386.321394} demonstrate that in standard iterative refinement, residuals computed in extra precision (typically twice the working precision) can guarantee the convergence of the relative error to the level of the working precision if the matrix $A$ has a condition number safely less than the reciprocal of the unit-roundoff, $u$. More recently, \cite{doi:10.1137/17M1122918} found that it is possible to produce solutions to iterative refinement with normwise relative error of order $u$ for systems with condition numbers of order $u^{-1}$ or larger by using a GMRES-based approach called GMRES-IR (see Algorithm \ref{alg:mix_gmres_ir}). The authors extended this work in \cite{doi:10.1137/17M1140819} to develop a general analysis of iterative refinement using up to three different precisions, as well as a general solver. This work shows that, depending on constraints on the condition number $\kappa(A)$, the most expensive part of the computation can often be carried out entirely in low precision without harming the accuracy.

However, to obtain theoretical results such as these, a great amount of effort is needed, and the analysis often relies on knowledge about the problem, e.g., the condition number, which may not be readily available in practice.  To circumvent these obstacles, one can switch to precision tuning tools that operate during compile time and runtime; see the survey \cite{10.1145/3381039} and references therein.

Another approach is to use an automated precision tuning method. Such methods, built upon the software level via static and dynamic analysis tools \citep{benkhalifa:hal-03978176},  can obtain suitable reduced-precision settings through trial and error.  The search-based tool Precimonious~\citep{6877460} uses high-performance dynamic programming approaches, which achieve a potentially lower precision setting for the program via Delta Debugging of the space of possible floating-point configurations within the target accuracy. Further, HiFPTuner \citep{10.1145/3213846.3213862} (High-level Floating-Point Tuner) operates at the source level to automatically tune the floating-point precision via profiling floating-point operations and estimating the numerical impact of the reduced-precision program. CRAFT \citep{10.1177/1094342016652462} builds reduced-precision configurations of floating-point programs based on the Dyninst binary instrumentation framework \citep{10.1177/109434200001400404}, and uses an automatic search process that identifies the precision-level sensitivity of various parts of an application. Search-based tools for finding a low-precision configuration typically require a dramatic increase in computational cost as the number of variables increases, and do not scale well to large programs. To further reduce the runtime of precision tuning, Blame Analysis~\citep{Rubio2016} is devoted to executing a program once, analyzing floating-point instructions to determine the minimum precision required for each operand, reducing the search space of Precimonious and improving scalability.  PyFloT \citep{9355251} automates mixed-precision tuning for HPC applications by leveraging runtime profiling, instruction-level granularity, and temporal locality to learn numerical behavior and recurring patterns in instruction execution. FPLearner \citep{10.1145/3597503.3623338} leverages a graph representation learning technique to predict the performance and accuracy of a mixed-precision program, which can then be used to accelerate dynamic precision tuning by reducing the number of program runs.

\begin{figure*}[h!]
\centering
\includegraphics[width=0.7\linewidth]{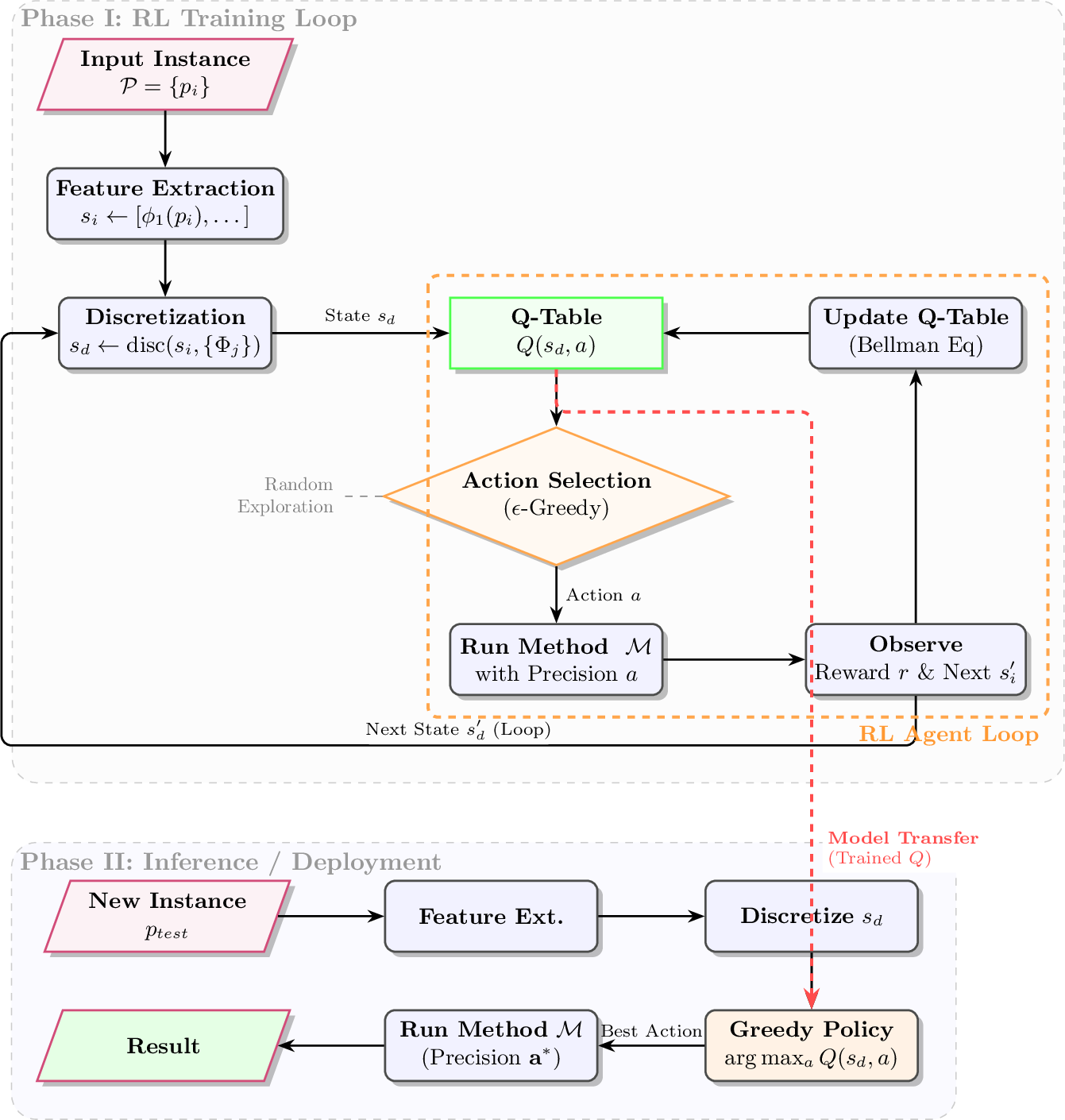}
\caption{Schematic of RL Framework: Phase I (top) depicts the iterative training loop where the agent updates the Q-table. Phase II (bottom) shows the inference workflow using the trained RL model.}
\label{fig:rl_complete_framework}
\end{figure*}

The aforementioned methods target floating-point programs, and their ability to handle unseen data remains unverified. Our proposed method operates in a different setting. It proceeds by modeling the rewards associated with multiple objectives and user-defined states (characteristics for the method to be defined in mixed-precision) and learning via several rounds of training on a specific, limited dataset. This approach can be generalized to a variety of unseen testing data. To the best of our knowledge, we are the first to present a work on precision autotuning for numerical algorithms that has verified generalization performance. In the following section, we will detail our framework and methodology for using RL to intelligently select precision configurations.

\section{A General RL Framework for Precision Selection}
\label{sec:general_framework}

We formalize the precision selection problem for a general algorithm $\mathcal{M}$ for solving linear systems as a contextual bandit, where an agent dynamically selects precision configurations for computational steps based on features of the problem instance. This framework is summarized in \figurename~\ref{fig:rl_complete_framework} and can be easily performed in an online learning routine. It intelligently optimizes the trade-off between numerical accuracy and computational efficiency, leveraging RL to adapt to diverse problem characteristics.
\subsection{Contextual Bandit Formulation}
\label{subsec:bandit_formulation}
Consider a numerical algorithm or method $\mathcal{M}$ designed to solve a problem instance $p \in \mathcal{P}$, where $\mathcal{P}$ denotes the set of problem instances (e.g., linear systems $Ax = b$). The method comprises $k$ computational steps, each requiring a precision choice from a finite set $\mathcal{A}_i = \{ u_{i,1}, u_{i,2}, \ldots, u_{i,m} \}$, for $i = 1, \ldots, k$, where $m = |\mathcal{A}_i|$ is the number of available precisions. The action space is the Cartesian product:
\begin{equation}
\mathcal{A} = \mathcal{A}_1 \times \mathcal{A}_2 \times \cdots \times \mathcal{A}_k,
\end{equation}
with cardinality $|\mathcal{A}| = m^k$. An action is a tuple $\mathbf{a} = (a_1, a_2, \ldots, a_k) \in \mathcal{A}$, where $a_i \in \mathcal{A}_i$ specifies the precision for the $i$-th computational step. To reduce the computational complexity and simplify the problem, we assume  $|\mathcal{A}_1|=|\mathcal{A}_2|=\ldots=|\mathcal{A}_k|$.

The contextual bandit is defined by:
\begin{itemize}
    \item \textbf{Context} ($s \in \mathcal{S}$): A feature vector describing the problem instance $p$, i.e., $s = [\phi_1(p), \phi_2(p), \ldots, \phi_d(p)] \in \mathbb{R}^d$, where $\phi_j: \mathcal{P} \to \mathbb{R}$ are features capturing properties such as condition number, sparsity, matrix norm, or matrix size.
    \item \textbf{Action} ($\mathbf{a} \in \mathcal{A}$): A precision configuration $\mathbf{a} = (a_1, \ldots, a_k)$, where $a_i$ determines the precision for the $i$-th step.
    \item \textbf{Reward} ($r(s, \mathbf{a})$): A scalar evaluating the performance of $\mathcal{M}$ on instance $p$ with action $\mathbf{a}$, defined as a weighted sum of $J$ objectives:
    \begin{equation}
    r(s, \mathbf{a}) = \sum_{j=1}^J w_j f_j(s, \mathbf{a}),
    \end{equation}
    where $f_j: \mathcal{S} \times \mathcal{A} \to \mathbb{R}$ are objective functions (e.g., accuracy, computational cost, convergence indicators), and $w_j > 0$ are weights balancing these objectives.
\end{itemize}

The goal is to learn a policy $\pi: \mathcal{S} \to \Delta(\mathcal{A})$, where $\Delta(\mathcal{A})$ is the probability simplex over $\mathcal{A}$ that maximizes the expected reward over a distribution of problem instances $\mathcal{D}$:
\begin{equation*}
J(\pi) = \mathbb{E}_{s \sim \mathcal{D}, \mathbf{a} \sim \pi(\cdot | s)} [ r(s, \mathbf{a}) ].
\end{equation*}

\subsection{Learning with Discretized Context Space}
\label{subsec:bandit_learning}

To learn the policy $\pi$ for determining precision choice, we employ a contextual bandit learning paradigm, discretizing the continuous state space $\mathcal{S}$ into a finite set to manage computational complexity. For each feature $\phi_j$ ($j = 1, \ldots, d$), we define bins $\Phi_j = \{ b_{j,1}, b_{j,2}, \ldots, b_{j,n_j} \}$, where $n_j$ is the number of bins (e.g., logarithmic bins for condition number and norm, linear bins for sparsity and diagonal dominance). The discretized state space is:
\begin{equation}\label{eq:discretized_state_space}
\mathcal{S}_d = \Phi_1 \times \Phi_2 \times \cdots \times \Phi_{d},
\end{equation}
with cardinality $|\mathcal{S}_d| = \prod_{j=1}^{d} n_j$. A state $s_d \in \mathcal{S}_d$ is obtained by mapping the continuous context $s = [\phi_1, \ldots, \phi_d]$ to a discrete index via binning:
\begin{equation}
s_d = \text{discretize}(s) = \left( \text{bin}(\phi_1(p)), \ldots, \text{bin}(\phi_d(p))) \right),
\end{equation}
where $\text{bin}(\cdot)$ assigns each feature to the nearest bin in $\Phi_j$, with clipping to ensure indices remain within bounds.

To handle the combinatorial action space $\mathcal{A} = \mathcal{A}_1 \times \cdots \times \mathcal{A}_k$
with cardinality $|\mathcal{A}| = m^k$, we directly learn a tabular action-value function
\[
Q : \mathcal{S}_d \times \mathcal{A} \to \mathbb{R},
\]
where $Q(s_d, \mathbf{a})$ estimates the expected reward obtained by applying the
precision configuration $\mathbf{a} = (a_1, \ldots, a_k) \in \mathcal{A}$
to the solver under discretized state $s_d \in \mathcal{S}_d$.

This formulation treats each precision configuration as a distinct action,
capturing the joint effect of precision choices across all computational steps.
The resulting Q-table has size $|\mathcal{S}_d| \cdot |\mathcal{A}| = |\mathcal{S}_d| \cdot m^k$,
which can become prohibitively large as the number of steps $k$ or precision levels $m$ increases.

In the following subsection, we introduce a structured reduction of the action space
to mitigate this exponential growth while preserving numerically meaningful configurations.

During training, the policy selects actions using an epsilon-greedy strategy to balance exploration and exploitation. For each episode, given a discretized state $s_d \in \mathcal{S}$, the agent selects an action $\mathbf{a} \in \mathcal{A}$, where $\mathbf{a} = (u_1, u_2, \ldots, u_k)$ represents a tuple of precisions. The set $\mathcal{A}$ contains (valid) precision combinations. The probability of selecting an action $\mathbf{a}$ is given by:
\begin{equation}
\pi(\mathbf{a} | s_d) =
\begin{cases}
1 - \epsilon + \epsilon / |\mathcal{A}| & \text{if } \mathbf{a} = \arg\max_{\mathbf{a}' \in \mathcal{A}} Q(s_d, \mathbf{a}'), \\
\epsilon / |\mathcal{A}| & \text{otherwise},
\end{cases}
\end{equation}
where $\epsilon \in [0, 1]$ is the exploration parameter, initialized to $\epsilon_0 = 1.0$ and decayed as $\epsilon \leftarrow \max(\epsilon_{\text{min}}, 1.0 - t / T)$ with a minimum exploration rate $\epsilon_{\text{min}}$.

After selecting an action $\mathbf{a}$, the environment applies the mixed-precision algorithm to the linear system $A \mathbf{x} = \mathbf{b}$, yielding a reward $R(s_d, \mathbf{a})$ based on accuracy, precision usage, convergence, and iteration count. The Q-table is updated with a one-step:
\begin{equation}
Q(s_d, \mathbf{a}) \leftarrow
Q(s_d, \mathbf{a}) + \alpha \Big( R(s_d, \mathbf{a}) - Q(s_d, \mathbf{a}) \Big),
\end{equation}
where $\alpha \in (0, 1]$ is the learning rate. The Q-values are not necessarily clipped in the implementation, relying on the reward function to manage numerical stability.

During inference, the agent selects the action that maximizes the Q-value for the given state:
\begin{equation}
\mathbf{a}^* = \arg\max_{\mathbf{a}' \in \mathcal{A}} Q(s_d, \mathbf{a}').
\end{equation}

\begin{algorithm}
\caption{Contextual Bandit Framework for Precision Selection}
\label{alg:general_framework_corrected}
\begin{algorithmic}[1]
\State \textbf{Input:} Training instances $\mathcal{P}=\{p_1,\ldots,p_N\}$, numerical method $\mathcal{M}$, precision sets $\{\mathcal{A}_i\}_{i=1}^k$, reward function $r$, episodes $T$, learning-rate schedule $\alpha_t$, exploration rate $\epsilon$, bin sets $\{\Phi_j\}_{j=1}^d$
\State Construct the joint action space $\mathcal{A}=\mathcal{A}_1\times\cdots\times\mathcal{A}_k$
\State Initialize a tabular action-value estimator $Q(s_d,\mathbf{a})$ for all $s_d\in\mathcal{S}_d$ and $\mathbf{a}\in\mathcal{A}$
\State Initialize visit counts $N(s_d,\mathbf{a})\gets 0$ for all $s_d\in\mathcal{S}_d$ and $\mathbf{a}\in\mathcal{A}$

\For{episode $t=1$ to $T$}
    \For{each instance $p_i\in\mathcal{P}$}
        \State Compute context $s_i \gets [\phi_1(p_i),\ldots,\phi_d(p_i)]$
        \State Discretize context $s_d \gets \mathrm{discretize}(s_i,\{\Phi_j\})$
        \State Select action $\mathbf{a}\in\mathcal{A}$ using an $\epsilon$-greedy policy:
        \[
        \mathbf{a} \gets
        \begin{cases}
        \text{a uniformly random action in }\mathcal{A}, & \text{with prob. }\epsilon,\\
        \arg\max_{\mathbf{a}'\in\mathcal{A}} Q(s_d,\mathbf{a}'), & \text{with prob. }1-\epsilon
        \end{cases}
        \]
        \State Run $\mathcal{M}$ with precision configuration $\mathbf{a}$ on $p_i$
        \State Observe reward $r_i \gets r(s_d,\mathbf{a})$
        \State Update visit count: $N(s_d,\mathbf{a}) \gets N(s_d,\mathbf{a}) + 1$
        \State Set learning rate, e.g.,
        \[
        \alpha \gets \frac{1}{N(s_d,\mathbf{a})}
        \]
        \State Update the tabular action-value estimator:
        \[
        Q(s_d,\mathbf{a}) \gets Q(s_d,\mathbf{a}) + \alpha\big(r_i - Q(s_d,\mathbf{a})\big)
        \]
    \EndFor
    \State Update $\epsilon$
\EndFor

\State \textbf{Inference:} For a new instance $p$, compute $s=[\phi_1(p),\ldots,\phi_d(p)]$, discretize $s_d=\mathrm{discretize}(s,\{\Phi_j\})$, and select
\[
\mathbf{a}^*=\arg\max_{\mathbf{a}\in\mathcal{A}} Q(s_d,\mathbf{a}).
\]
\State Apply $\mathcal{M}$ with precision configuration $\mathbf{a}^*$
\end{algorithmic}
\end{algorithm}

Our learning framework for precision selection is formulated as a contextual bandit over a discretized context space. For each problem instance $p=(A,b)\in\mathcal{P}$, we compute a feature vector $s=[\phi_1(p),\ldots,\phi_d(p)]$ and map it to a discrete context $s_d\in\mathcal{S}_d$. The agent then selects a joint precision configuration $\mathbf{a}=(a_1,\ldots,a_k)\in\mathcal{A}$ for the $k$ computational steps of the solver. After executing the mixed-precision algorithm with action $\mathbf{a}$, the environment returns a scalar reward $r(s_d,\mathbf{a})$ that balances numerical accuracy, computational cost, convergence behavior, and iteration count.

To estimate the expected reward of each state-action pair, we maintain a single tabular function $Q(s_d,\mathbf{a})$. During training, actions are sampled using an $\epsilon$-greedy exploration rule, and the table is updated using the incremental estimator
\[
Q(s_d,\mathbf{a})\leftarrow Q(s_d,\mathbf{a})+\alpha_t(s_d,\mathbf{a})\big(r(s_d,\mathbf{a})-Q(s_d,\mathbf{a})\big).
\]

In the inference phase, for a new instance $p$, the context $s$ is computed and discretized to $s_d$, and the precision configuration is selected greedily as
\[
\mathbf{a}^*(s_d)=\arg\max_{\mathbf{a}\in\mathcal{A}} Q(s_d,\mathbf{a}).
\]
Using a single Q-table over the reduced joint action space preserves the interaction among precision choices across different computational steps while keeping the learning problem tractable. The feature-based context enables the framework to adaptively select numerically effective and computationally efficient precision configurations, as validated in the numerical experiments in Section~\ref{sec:exps}.

Our RL learning framework optimizes both arithmetic efficiency (by favoring lower-precision steps) and solver performance (e.g., convergence and accuracy). In Section~
\ref{sec:gmres_application}, we will apply this framework to GMRES-IR, and detail how we model the reward function based on the earnings and cost for the steps following low precision.

\paragraph{Theoretical Justification of Discretization}
Our contextual bandit framework relies on the discretization of the continuous context space $\mathcal{S}$ into a finite set $\mathcal{S}_d$. This raises the question of how much optimality is lost due to this approximation. We provide a bound below showing that, under a mild smoothness assumption on the expected reward, the performance loss induced by discretization is controlled by the granularity of the bins.

\begin{assumption}[Lipschitz continuity of the expected reward]
\label{ass:lipschitz}
Let $\mu(s,\mathbf{a}) := \mathbb{E}[r(s,\mathbf{a})]$ denote the expected reward.
We assume, for all $\mathbf{a} \in \mathcal{A}$, there exists a constant $L > 0$ such that
\begin{equation}
\label{eq:lipschitz}
|\mu(s,\mathbf{a}) - \mu(s',\mathbf{a})| \le L \| s - s' \|, \quad \forall s, s' \in \mathcal{S}.
\end{equation}
\end{assumption}

\begin{proposition}[Discretization error bound]
\label{prop:discretization}
Let $s \in \mathcal{S}$ be a continuous context, and let $s_d \in \mathcal{S}_d$ be its discretized representation.
Assume that each discretization bin has diameter at most $\Delta$, i.e.,
\[
\| s - \omega(s_d) \| \le \Delta,
\]
where $\omega(s_d)$ denotes a representative point (e.g., bin center) of $s_d$.

Define the optimal action in the continuous space and the discretized policy, respectively, as
\begin{equation*}
\begin{aligned}
\mathbf{a}^*(s) &= \arg\max_{\mathbf{a} \in \mathcal{A}} \mu(s,\mathbf{a}), \\ \mathbf{a}_d^*(s_d) &= \arg\max_{\mathbf{a} \in \mathcal{A}} \mu(\omega(s_d), \mathbf{a}).
\end{aligned}
\end{equation*}

Under Assumption~\ref{ass:lipschitz}, the performance loss due to discretization is bounded by
\begin{equation}
\label{eq:discretization_bound}
\mu(s, \mathbf{a}^*(s)) - \mu(s, \mathbf{a}_d^*(s_d)) \le 2 L \Delta.
\end{equation}
\end{proposition}

\begin{proof}
Let $\omega = \omega(s_d)$ be the representative point associated with the discretized state $s_d$.
By Assumption~\ref{ass:lipschitz}, for any $\mathbf{a} \in \mathcal{A}$, we have
\begin{equation}
\label{eq:lipschitz_step}
|\mu(s,\mathbf{a}) - \mu(\omega,\mathbf{a})| \le L \| s - \omega \| \le L \Delta.
\end{equation}

By definition of $\mathbf{a}^*(s)$,
\[
\mu(s,\mathbf{a}^*(s)) \le \mu(\omega,\mathbf{a}^*(s)) + L\Delta.
\]

Since $\mathbf{a}_d^*(s_d)$ maximizes $\mu(\omega,\mathbf{a})$ over $\mathcal{A}$, we have
\[
\mu(\omega,\mathbf{a}^*(s)) \le \mu(\omega,\mathbf{a}_d^*(s_d)).
\]

Applying the Lipschitz bound again,
\[
\mu(\omega,\mathbf{a}_d^*(s_d)) \le \mu(s,\mathbf{a}_d^*(s_d)) + L\Delta.
\]

Combining the above inequalities yields
\[
\mu(s,\mathbf{a}^*(s)) \le \mu(s,\mathbf{a}_d^*(s_d)) + 2L\Delta,
\]
which proves the result.
\end{proof}

Proposition~\ref{prop:discretization} shows that the loss induced by discretizing the context space is controlled by the bin diameter $\Delta$. In particular, if the expected reward varies smoothly with respect to problem features (such as condition number and matrix norm), then finer discretizations lead to policies that approach the optimal continuous-context policy. This provides a theoretical justification for the discretization strategy used in our contextual bandit framework.

\paragraph{Action Space Reduction}
\label{subsec:greedy_strategy}

If the set of precisions is too large, then the size of the  action space will exponentially increase, which significantly increases the training load.

Let $\mathcal{A}_1 = \{ u_1, u_2, \ldots, u_m \}$ be the set of precisions ordered by increasing significand bits. For the solver $\mathcal{M}$ and $k$ computational steps for which precisions must be selected, the action space $\mathcal{A} = \mathcal{A}_1^k$ has cardinality $m^k$, where $m$ is the number of precisions available to be chosen. To manage this exponential growth, we prune the action space by imposing a precision order constraint and limiting to the top-$k_{\text{top}}$ combinations.

Here, we write $u_j < u_{j'}$ if $u_j$ has fewer significand bits than $u_{j'}$. We enforce that the selected precisions for the $k$ computational steps satisfy
\begin{equation}
\label{eq:precision_constraint}
u'_1 \leq u'_2 \leq u'_3 \leq \ldots  \leq u'_k, \qquad u'_j \in \mathcal{A}_1
\end{equation}

This can naturally lead to valid precision combinations for some algorithms that is meaningful, as discussed in \cite{vi22}.
This idea is similar to modified delta-debugging in \cite{6877460}, which uses a greedy search approach instead of seeking a global minimum; it might miss potential reduced-precision configurations that suit the algorithm. The principle also naturally enables early stopping. For example, if precision $u'_i$ fails, then it is not necessary to try a lower precision in step $i$ or to try lower precision for steps $j > i$ if it is already employed. Another benefit is that it allows users to define the precision selection for the $k$ computational steps in terms of their importance (the most important one is assigned to the last step $k$). One benefit of using this strategy is to reduce the action space to
\begin{equation}
\label{eq:reduced_action_space}
|\mathcal{A}_{\text{reduced}}| = \binom{m + k - 1}{k} = \frac{(m+k-1)!}{k!(m-1)!}.
\end{equation}

In the following experiments, we prune the action space from 256 to 35, a reduction of approximately 86\%.

The exploration strategy is epsilon-greedy with linear decay. For episode $t$, the probability of selecting a random action from $\mathcal{A}_{\text{reduced}}$ is
\begin{equation}
\epsilon_t = \max(\epsilon_{\text{min}}, 1.0 - t / T).
\end{equation}
This ensures exploration of lower-precision configurations early in training, shifting to exploitation of the best-known actions as $\epsilon_t$ decreases. The reduced action space $\mathcal{A}_{\text{reduced}}$ focuses on numerically stable configurations, accelerating convergence of the RL agent for the solver.

\section{Case Study: GMRES-Based Iterative Refinement}
\label{sec:gmres_application}

Iterative numerical methods, such as those for solving linear systems $Ax = b$, rely on a careful choice of arithmetic precisions to balance computational efficiency and numerical accuracy. The choice of precision for each computational step (e.g., working precision, residual computation) significantly impacts performance, yet optimal selection is challenging due to the complex dependence on system properties, such as condition number and sparsity. Traditional approaches often employ fixed or heuristic precision settings, which may lead to suboptimal performance across diverse problem instances. RL provides a data-driven approach to adaptively select precisions based on system characteristics, optimizing multiple objectives, such as residual norm and iteration count.

We instantiate the general RL framework for the GMRES-IR method \citep{doi:10.1137/17M1140819}, selecting precisions for four computational steps to solve linear systems $Ax = b$, reconciling the competing demands of numerical accuracy and computational efficiency. Consider a linear system $Ax = b$, where $A \in \mathbb{R}^{n \times n}$ is non-singular, and $b \in \mathbb{R}^n$. GMRES-IR iteratively refines an initial solution $x_0$ as:
\begin{equation*}
x_{i+1} = x_i + z_i, \quad \text{where} \quad Az_i = r_i,
\end{equation*}
and $r_i = b-Ax_i$. The system $Az_i=r_i$ can be solved via preconditioned GMRES, i.e., we solve $M^{-1}Az_i = M^{-1}r_i$, where $M \approx A$ is a preconditioner. In this study, we use the preconditioner $M =LU$ (typically performed by backward and forward substitution) via LU factorization.  The algorithm is presented in Algorithm~\ref{alg:mix_gmres_ir}. We refer to the loop in  line~3 as \emph{outer iterations} and the iterations within the GMRES solve in line~5 as \emph{inner iterations}.

Our mixed-precision algorithms focus on four computational steps, each requiring a precision choice:
\begin{enumerate}
    \item \textbf{LU Factorization and Initial Solution}: Compute $M = LU \approx A$ and solve $M x_0 = b$ in precision $u_f$,
    \item \textbf{Residual Computation}: Compute the residual $r_i = b - Ax_i$ in precision $u_r$.
    \item \textbf{Working Precision for GMRES}: Solve $M^{-1} A z_i = M^{-1} r_i$ using GMRES in precision $u_g$.
    \item \textbf{Solution Update}: Compute $x_{i+1} = x_i + z_i$ in precision $u$.
\end{enumerate}

\begin{algorithm}
\caption{Mixed-precision GMRES-IR}\label{alg:mix_gmres_ir}
\begin{algorithmic}[1]
\Require an $n \times n$ matrix $A$ and a right-hand side $b$; $u_f$ is precision for factorization;
\Ensure Approximate solution $\hat{x}$ to $A\hat{x} = b$.

\State Compute the LU factorization $A = LU$. \hfill \textcolor{tomato}{$u_f$}
\State Initialize $x_0$ (to, e.g., $x_0=U^{-1}L^{-1}(b)$). \hfill \textcolor{tomato}{$u_f$}
\For{$i = 0, \dots$, converged}
    \State Compute $r_i = b - A x_i$. \hfill \textcolor{darkblue}{$u_r$}
    \State Solve $U^{-1}L^{-1}A z_i = U^{-1}L^{-1}r_i$ using GMRES.  \hfill
    \textcolor{darkgreen}{$u_g$}
    \State Compute $x_{i+1} = x_i + z_i$. \hfill \textcolor{orange}{$u$}
\EndFor
\end{algorithmic}
\end{algorithm}

It is worth noting that mixed precision can also be applied directly within the GMRES algorithm itself (see, for example, \cite{vi22}). In this work, however, we limit our attention to GMRES implemented with a single, consistent precision for simplicity.

The goal is to use RL to select an action $\mathbf{a} = (u_f, u, u_g, u_r)$ to minimize the final residual norm $\|b - Ax_{\text{solve}}\|_2$ and the total number of iterations $T$, subject to a trade-off between $T_{\text{max}}$ and the convergence tolerance $\tau$.

Inspired by stopping criteria from \cite{10.1145/1141885.1141894}, we consider the algorithm converged if:
\begin{enumerate}
    \item \textbf{Convergence}: The solver converges if the relative update norm is sufficiently small:
    \begin{equation}
    \label{eq:stop_convergence}
    \frac{\|z_i\|_\infty}{\|x_i\|_\infty} \leq u_{\text{work}},
    \end{equation}
    where $u_{\text{work}}$ is the unit roundoff for working precision, ensuring the update is on the order of the highest precision's roundoff error.
    \item \textbf{Stagnation}: The solver stops if the update norms indicate insufficient progress:
    \begin{equation}
    \label{eq:stop_stagnation}
    \frac{\|z_i\|_\infty}{\|z_{i-1}\|_\infty} \geq \tau,
    \end{equation}
    where $\tau > 0$ is the stagnation tolerance, indicating that the updates are not changing significantly.
    \item \textbf{Maximum Iterations}: The solver stops if the number of iterations reaches a predefined limit:
    \begin{equation}
    \label{eq:stop_max_iter}
    i \geq i_{\text{max}},
    \end{equation}
    where $i_{\text{max}}$ is the maximum number of iterations.
\end{enumerate}

To evaluate the accuracy and stability of the linear solve, we define the normwise relative forward error, and normwise relative backward error:
\begin{equation}\label{eq:rela_err}
\begin{aligned}
\texttt{ferr} &=
\frac{\|x_{\text{solve}} - x_{\text{true}}\|_\infty}{\|x_{\text{true}}\|_\infty}, \\
\texttt{nbe} &=
\frac{\|b - A x_{\text{solve}}\|_\infty}
{\|A\|_\infty \|x_{\text{solve}}\|_\infty + \|b\|_\infty}.
\end{aligned}
\end{equation}

\subsection{Contextual Bandit Formulation for Precision selection}
\label{subsec:gmres_bandit}

The contextual bandit for the iterative refinement solver with preconditioned GMRES is defined by specifying the context, action, and reward components, tailored to optimize precision selection for solving linear systems $Ax = b$. The context, denoted $s \in \mathbb{R}^d$, where $d=2$, captures properties of the linear system and is given by:
\begin{equation}
s = [\log_{10}(\max(\kappa(A), \delta_c)), \log_{10}(\max(\|A\|_\infty, \delta_n))],
\end{equation}
where $\kappa(A)$ is the condition-number of $A$ that can be approximated via an efficient algorithm (e.g., Hager-Higham \cite{Hager:1984:CE, Higham:1987:SCN} and GNNs \cite{carson26estcond}), and $\delta_c > 0$  ensures numerical stability in logarithmic scaling, $\|A\|_\infty = \max_i \sum_j |a_{ij}|$ is the infinity norm, and $\delta_n > 0$ prevents numerical issues. These features characterize the numerical stability and scaling of the linear system and can be made inexpensive through approximate feature extraction.

To learn the policy $\pi$, we employ a contextual bandit, discretizing the continuous state space $\mathcal{S} \subset \mathbb{R}^d$ into a finite set to manage computational complexity. For each feature $\phi_j$, $j = 1, \ldots, d$, we define bins $\Phi_j$. The continuous state space $\mathcal{S} \subset \mathbb{R}^d$ is discretized into a finite set $\mathcal{S}_d$, as in \eqref{eq:discretized_state_space}. For features $\phi_1 = \log_{10}(\max(\kappa(A), \delta_c))$ and $\phi_2 = \log_{10}(\max(\|A\|_\infty, \delta_n))$, we define bins:
\begin{itemize}
    \item $\Phi_1 = \log([\phi_{1,\text{min}}, \phi_{1,\text{max}}])$, divided into $n_1$ bins for the condition number (e.g., use approximate 1-norm).
    \item $\Phi_2 = \log([\phi_{2,\text{min}}, \phi_{2,\text{max}}])$, divided into $n_2$ bins for the matrix norm.
\end{itemize}
The discretized state space is $\mathcal{S}_d = \Phi_1 \times \Phi_2$, with cardinality $|\mathcal{S}_d| = n_1 \cdot n_2$. A state $s_d \in \mathcal{S}_d$ is computed as:
\begin{equation}
s_d = \text{discretize}(s) = \left( \text{bin}(\phi_1(s)), \text{bin}(\phi_2(s)) \right),
\end{equation}
where $\text{bin}(\phi_j(s))$ assigns $\phi_j(s)$ to the nearest bin in $\Phi_j$, clipped to $[0, |\Phi_j|-1]$. The state index is:
\begin{equation}
s_d = \text{bin}(\phi_1) \cdot n_2 + \text{bin}(\phi_2).
\end{equation}

The action $\mathbf{a} \in \mathcal{A}$ is a tuple $\mathbf{a} = (u_f, u, u_g, u_r), \quad u_i \in \mathcal{A}_1$ where $\mathcal{A}_1 = \{ u_1, \ldots, u_m \}$ is the set of precisions, and $m = |\mathcal{A}_1|$. Following \eqref{eq:reduced_action_space}, the action space $\mathcal{A}_{\text{reduced}} \subseteq \mathcal{A}$ is the top $k$ elements of $\mathcal{A} = \mathcal{A}_1^4$ is constrained by $u_f \leq u \leq u_g \leq u_r$ where $\leq$ denotes ordering by increasing significand bits.

The reward $r(s, \mathbf{a})$ is designed to lead RL to choose precision configurations that lead the linear solvers to have an acceptable accuracy and number of iterations until convergence. \rev{We use the general form}
\rev{
\begin{equation}
\label{eq:reward}
\begin{aligned}
R(s_d, \mathbf{a}) =
{}& w_2 f_{\text{precision}}(s_d, \mathbf{a}) \\
&+ w_1 f_{\text{accuracy}}(s_d, \mathbf{a}) \\
&- \lambda_{\text{iter}}
\log_2(\max(T_{\mathrm{GMRES}},1)),
\end{aligned}
\end{equation}
}
\rev{where the weights $w_1, w_2 > 0$ balance accuracy and cost,
$T_{\mathrm{GMRES}}$ is the total number of inner GMRES iterations accumulated over all outer refinement steps for the solve, and $\lambda_{\text{iter}}\geq 0$ controls the strength of the iteration penalty.} The individual components are defined as follows:
\begin{itemize}
\item \textbf{Precision term}: $f_{\text{precision}}(s_d, \mathbf{a})$ encourages the use of lower-precision formats based on their significand bits. For each precision $p \in \{u_f, u, u_g, u_r\}$, the cost is weighted by the ratio of significand bits relative to FP64:
    \begin{equation}
    \begin{aligned}
    f_{\text{precision}}(s_d, \mathbf{a})
    ={}& \sum_{p \in \{u_f, u, u_g, u_r\}}
    q_p,\\
    q_p
    ={}& \frac{t_{\text{FP64}}}{t_p d_{\kappa}},\\
    d_{\kappa}
    ={}& 1+\log_{10}(\max(\kappa, 1)),
    \end{aligned}
    \end{equation}
    where $t_u$ is the number of significand bits for precision $u$, and $\kappa$ is the condition number of the matrix $\mathbf{A}$.
\xy{This bit-count proxy is intentionally architecture independent. If platform-specific timings, memory traffic, or energy measurements are available, the same term can be replaced by a weighted version with stage-dependent coefficients for $u_f$, $u$, $u_g$, and $u_r$. We keep the learning reward independent of a particular architecture; the separate solver-level CPU validation then measures the resulting policy using compiled native FP16, FP32, and FP64 arithmetic.}
\item \textbf{Accuracy term}: $f_{\text{accuracy}}(s_d, \mathbf{a})$ penalizes loss of solution accuracy and residual quality. \rev{In the released implementation, the training reward uses two quantities: a reference-solution relative error and a scaled forward-error/residual-quality proxy. Let $x_{\mathrm{ref}}$ denote the known or FP64 reference solution used during training, and let $\eta$ denote the final forward-error/residual-quality scalar returned by the solver implementation for the sampled precision action. We define}
\rev{
\begin{equation}
\label{eq:reward_error_terms}
\begin{aligned}
e_{\mathrm{ref}} &=
\frac{\|x_{\mathrm{solve}}-x_{\mathrm{ref}}\|_{\infty}}
{\|x_{\mathrm{ref}}\|_{\infty}+\varepsilon},\\
e_{\mathrm{scaled}} &=
\frac{\eta}
{\|A\|_{\infty}\|x_{\mathrm{ref}}\|_{\infty}+\|b\|_2+\varepsilon}.
\end{aligned}
\end{equation}
}
\rev{The reported \texttt{ferr} and \texttt{nbe} in \eqref{eq:rela_err} remain the evaluation metrics used in the tables and discussion; in particular, \texttt{nbe} is not directly one of the logarithmic training-reward terms in the current implementation. The implemented accuracy reward is}
\rev{
    \begin{equation}
    \label{eq:reward_accuracy}
    \begin{aligned}
    f_{\text{accuracy}}(s_d, \mathbf{a})
    &=
    \begin{cases}
    -5C_1,
    & E>1,\\
    -C_1 L,
    & E\leq 1,
    \end{cases}\\
    E&=\max(e_{\mathrm{ref}},e_{\mathrm{scaled}}),\\
    L&=\ell_{\mathrm{ref}}+\ell_{\mathrm{scaled}},\\
    \ell_{\mathrm{ref}}
    &=\log_{10}(\max(e_{\mathrm{ref}},\varepsilon)),\\
    \ell_{\mathrm{scaled}}
    &=\log_{10}(\max(e_{\mathrm{scaled}},\varepsilon)).
    \end{aligned}
    \end{equation}
}
    \rev{where $\varepsilon$ is a small constant (we use $10^{-10}$). The first branch is a hard cap for large-error actions.}

\item \textbf{Penalty}: \rev{The last term in \eqref{eq:reward} penalizes computational cost through the accumulated inner GMRES iteration count. We use the logarithm because raw inner-iteration counts can vary enough to dominate the accuracy and precision terms; the logarithmic scale preserves the ordering of more expensive solves while keeping the reward components on comparable scales. In the reported dense experiments, the tabulated average GMRES counts are mostly between about $2$ and $21$, corresponding to an unweighted logarithmic penalty of roughly $1$ to $4.4$. The sparse and PDE-generated experiments use the same form with a smaller coefficient, $\lambda_{\text{iter}}=0.5$, to avoid over-penalizing Krylov work on harder sparse systems, while the dense experiments use $\lambda_{\text{iter}}=1$. Convergence diagnostics and failure flags are used for termination, truncation, logging, and evaluation, but the released reward implementation does not add a separate convergence-bonus term. The precision term is typically of order $1$--$10$ before multiplication by $w_2$, and the accuracy term is of comparable order for the observed training errors. These constants were chosen to distinguish a conservative policy ($W_1$) from a more aggressive low-precision policy ($W_2$), rather than tuned to a particular hardware platform.}
\end{itemize}

We use a contextual bandit update for a single Q-table $Q: \mathcal{S}_d \times \mathcal{A} \to \mathbb{R}$. Actions are selected using an epsilon-greedy strategy with
\begin{equation}
\epsilon_t = \max(\epsilon_{\text{min}}, 1.0 - t / T),
\end{equation}
as in Section~\ref{subsec:bandit_learning}. The Q-table is updated as
\begin{equation}
\begin{aligned}
Q(s_d, \mathbf{a}) \leftarrow
\, Q(s_d, \mathbf{a}) + \alpha \Big( R(s_d, &\mathbf{a}) - Q(s_d, \mathbf{a}) \Big), \\
&\mathbf{a} \in \mathcal{A}_{\text{reduced}}.
\end{aligned}
\end{equation}

The precision selection RL framework specialized for GMRES-IR is presented in Algorithm~\ref{alg:gmres_precision}. It applies RL to a training set of $N$ linear systems, selecting precisions via an epsilon-greedy strategy and updating the Q-table based on solver performance. The GMRES method is adapted to handle mixed-precision computations, with the preconditioner applied in precision $u_g$.

\begin{algorithm}
\caption{Contextual Bandit Learning for GMRES-IR Precision Selection}
\label{alg:gmres_precision}
\begin{algorithmic}[1]
\State \textbf{Input:} Training systems $\{ (A_j, b_j) \}_{j=1}^N$, GMRES-IR method $\mathcal{M}$, precision set $\mathcal{A}_1 = \{ u_1, \ldots, u_m \}$, reward function $r$, episodes $T$, tolerance $\tau$, max iterations $T_{\text{max}}$, learning rate $\alpha$, minimum exploration probability $\epsilon_{\text{min}}$, bins $\{ \Phi_1, \Phi_2 \}$, top-$k$ parameter $k_{\text{top}}$
\State Compute $\mathcal{A}_{\text{reduced}}$ in terms of $\mathcal{A} = \mathcal{A}_1^4$ with $k_{\text{top}}$ elements satisfying \eqref{eq:precision_constraint}
\State Initialize tabular action-value estimator $Q(s_d, \mathbf{a})$ for $s_d \in \mathcal{S}_d$, $\mathbf{a} \in \mathcal{A}_{\text{reduced}}$
\State Initialize environment $\mathcal{E}$ with systems $\{ (A_j, b_j) \}$
\For{episode $t = 1$ to $T$}
    \For{system $(A_j, b_j)$, $j = 1, \ldots, N$}
        \State Compute context $s_j = [\log_{10}(\max(\kappa(A_j), \delta_c)), \log_{10}(\max(\|A_j\|_\infty, \delta_n))]$
        \State Discretize state $s_d = \text{discretize}(s_j, \{ \Phi_1, \Phi_2 \})$
        \State Compute $\epsilon_t = \max(\epsilon_{\text{min}}, 1.0 - t / T)$
        \State Select action $\mathbf{a} = (u_f, u, u_g, u_r) \in \mathcal{A}_{\text{reduced}}$ via:
        \begin{equation*}
        \mathbf{a} =
        \begin{cases}
        \text{random choice from } \mathcal{A}_{\text{reduced}}, & \text{if } \text{with } \epsilon_t \text{ probability}, \\
        \arg\max_{\mathbf{a}' \in \mathcal{A}_{\text{reduced}}} Q(s_d, \mathbf{a}'), & \text{otherwise}.
        \end{cases}
        \end{equation*}
        \State Solve $A_j x = b_j$ with $\mathcal{M}$ using $\mathbf{a}$:
        \State \quad Compute $M = LU \approx A_j$ in precision $u_f$
        \State \quad Solve $M x_0 = b_j$ in precision $u_f$
        \State \quad \textbf{for} $l = 0$ to $T_{\text{max}} - 1$:
        \State \quad \quad Compute $r_l = b_j - A_j x_l$ in precision $u_r$
        \State \quad \quad Solve $M^{-1} A_j z_l = M^{-1} r_l$ using GMRES in precision $u_g$
        \State \quad \quad Update $x_{l+1} = x_l + z_l$ in precision $u$
        \State \quad \quad \textbf{if} stopping criteria is met or failure occurs: \textbf{break}
        \State \quad Compute reward $r(s_j, \mathbf{a})$ per \eqref{eq:reward}
        \State Update tabular action-value estimator:
        \begin{equation*}
        Q(s_d, \mathbf{a}) \leftarrow
        Q(s_d, \mathbf{a}) + \alpha \Big( R(s_d, \mathbf{a}) - Q(s_d, \mathbf{a}) \Big)
        \end{equation*}
    \EndFor
\EndFor
\State \textbf{Inference:} For new system $(A, b)$, compute $s = [\log_{10}(\max(\kappa(A), \delta_c)), \log_{10}(\max(\|A\|_\infty, \delta_n))]$, discretize $s_d$, select $\mathbf{a} = \arg\max_{\mathbf{a} \in \mathcal{A}_{\text{reduced}}} Q(s_d, \mathbf{a})$, solve with $\mathcal{M}$ using $\mathbf{a}$
\end{algorithmic}
\end{algorithm}

\section{Experiments}\label{sec:exps}

We conducted experiments to evaluate the proposed RL framework for precision selection in GMRES-IR for solving linear systems $p = (A, b) \in \mathcal{P}$, where $A \in \mathbb{R}^{n \times n}$ is a non-singular matrix and $b \in \mathbb{R}^n$. The systems are derived from synthetically generated and PDE-discretized matrices. \xy{The reported dense, sparse, and PDE experiments use the standard precision set $\{\text{BF16},\text{TF32},\text{FP32},\text{FP64}\}$. The implementation includes PyTorch/native casting pathways for standard formats where available. Thus the framework can support FP8-like or custom exponent/significand formats by adding them to the action space, but such custom formats are not included in the reported experimental results. The reported solver-level CPU timings use a separate native C++ GMRES-IR implementation with FP16, FP32, and FP64.}

We generated problems $Ax = b$ using three settings, each designed to evaluate the RL-driven precision selection for the GMRES-IR solver under different numerical conditions versus a full double precision routine (in the following, we use FP64 and double precision interchangeably).  The agent with learning rate $\alpha$ of $0.5$ is trained within a custom Gym environment \citep{brockman2016openai} with 100 episodes, which simulates the process of solving a linear system using the GMRES-IR solver. To reduce the action space and accelerate computation, we adopted the reduction strategy outlined in Section~\ref{subsec:greedy_strategy}; after enforcing the precision-order constraint, the reported scripts use the top $k_{\text{top}}=25$ valid precision combinations for the four controlled stages.

\xy{The experiments are implemented with deterministic random seeds. For each of the dense, randomly sparse, and PDE settings, the scripts use $N_{\text{train}}=100$ and $N_{\text{test}}=100$, learning rate $\alpha=0.5$, $\epsilon_{\min}=0.1$, 10 logarithmic bins for the condition-number feature, 10 logarithmic bins for the matrix-norm feature, and 100 training episodes. The reported scripts set the maximum outer-iteration budget to 9999, use one restarted inner GMRES cycle per outer refinement step, and stop after three consecutive stagnating outer updates. The native CPU validation trains on 50 dense systems from the low- and medium-condition regimes and validates on 100 dense systems spanning low, medium, and high regimes, with matrix sizes 1000--1500. The FP64 baseline uses the same tolerance as the RL-guided solver.}


In the following experiments, we evaluate the resulting relative forward errors (as defined in \eqref{eq:rela_err}), the number of iterative refinement iterations required for convergence, and the total number of inner GMRES iterations required for convergence. \xy{The quantity reported as ``Avg. GMRES iter.'' in the tables is the total number of inner GMRES iterations accumulated over all outer iterative-refinement steps for one solve, averaged over the corresponding test subset.} All the computed results for the relative forward and backward errors (abbreviated as error in the tables and figures for simplicity in the following) are preserved to 2 significant digits. For a fair comparison, we set the same convergence tolerance $\tau$ for both RL baseline and the reference baseline.

\subsection{Problem Setup}

To ensure sufficient diversity in the data, we synthetically generated the linear systems, comprising $N_{\text{train}}$ training and $N_{\text{test}}$ testing systems $\{ \mathcal{P}_i = (A_i, b_i) \}_{i=1}^{N_{\text{train}} + N_{\text{test}}}$. The data generation iterated until $N_{\text{train}} + N_{\text{test}} = 200$ systems were obtained. Specifically, we set $N_{\text{train}} = 100$ training and $N_{\text{test}} = 100$ testing systems. Across the synthetic dense, random sparse, and PDE benchmarks, the generation process incorporates randomized target matrix sizes between 100 and 500 and prescribed or coefficient-induced condition-number ranges; for the PDE benchmark, square-grid discretizations round the realized matrix sizes up to 529. This diversity is critical for evaluating our RL-based precision selection strategy. We use precision set $\mathcal{U} = \{ \text{BF16}, \text{TF32}, \text{FP32}, \text{FP64} \}$ and  the discretized state as in \eqref{eq:discretized_state_space}, based on two features: $\phi_1 = \log_{10}(\max(\kappa(A), \delta_c))$ (condition number) and $\phi_2 = \log_{10}(\max(\|A\|_\infty, \delta_n))$ (matrix norm), where $\delta_c, \delta_n > 0$ ensure numerical stability. These features $\Phi_1$ and $\Phi_2$ are separately binned into 10 bins ($n_1 = n_2 = 10$), respectively and their minimum and maximum value in terms of the training set.  The discretized state space is $\mathcal{S}_d = \Phi_1 \times \Phi_2$, with cardinality $|\mathcal{S}_d| = n_1 \cdot n_2$, and states are indexed as in \eqref{eq:discretized_state_space}.

\xy{In the reported experiments, the condition-number feature is taken from the corresponding data-generation or preprocessing routine. For the dense synthetic systems, the generated dense matrix is used to compute $\kappa(A)$ with a dense condition-number calculation. For the randomly sparse positive-definite systems, the same dense calculation is applied after converting the generated sparse matrix to a dense array. For the PDE-generated systems, \texttt{cond\_method=auto} evaluates a dense condition number for the current 100--500 matrix-size range, while larger matrices, or an explicit \texttt{cond\_method=onenormest} setting, use a Hager--Higham-type one-norm estimator implemented through a sparse LU factorization and repeated triangular solves for estimating $\|A^{-1}\|_1$ \citep{Hager:1984:CE, Higham:1987:SCN}. In the native CPU validation, the controlled target condition number is used only to generate and report the low/medium/high regimes; the RL state uses a Hager--Higham-type one-norm condition estimate computed from each generated matrix. The reported CPU speedups therefore isolate native GMRES-IR execution after feature construction, rather than full deployment timing that includes condition-estimation overhead.}


To evaluate the performance of the RL-based solver $\mathcal{M}$, we define the success rate across a test set of linear systems $\{ \mathcal{P}_1, \ldots, \mathcal{P}_M \} \subset \mathcal{S}$. For each system $\mathcal{P} = (A, b)$, the solver produces a solution $x_{\text{rl}}$ using the learned policy $\pi$.

The performance is assessed using the maximum of the two error metrics \texttt{ferr} and \texttt{nbe}, defined by:
\begin{equation*}
\epsilon_{\text{max}}(\mathcal{P}, \mathbf{a}) = \max(\texttt{ferr}, \texttt{nbe}).
\end{equation*}

\begin{figure}[htp!]
    \centering
    \includegraphics[width=0.99\linewidth]{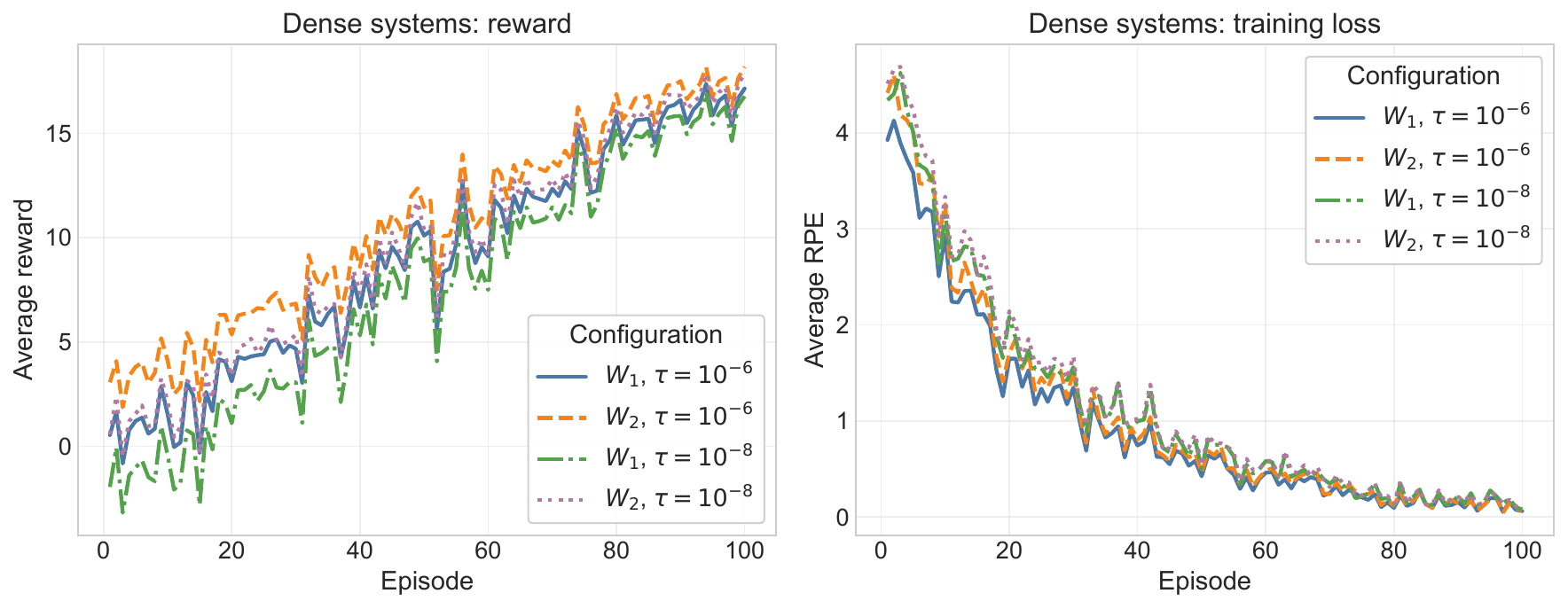}
    \caption{Episode rewards and training loss/RPE for the dense systems across the four reward-weight/tolerance settings: $(W_1, \tau=10^{-6})$, $(W_2, \tau=10^{-6})$, $(W_1, \tau=10^{-8})$, and $(W_2, \tau=10^{-8})$.}
    \label{fig:dense_training}
\end{figure}

\begin{figure}[htp!]
    \centering
    \includegraphics[width=0.99\linewidth]{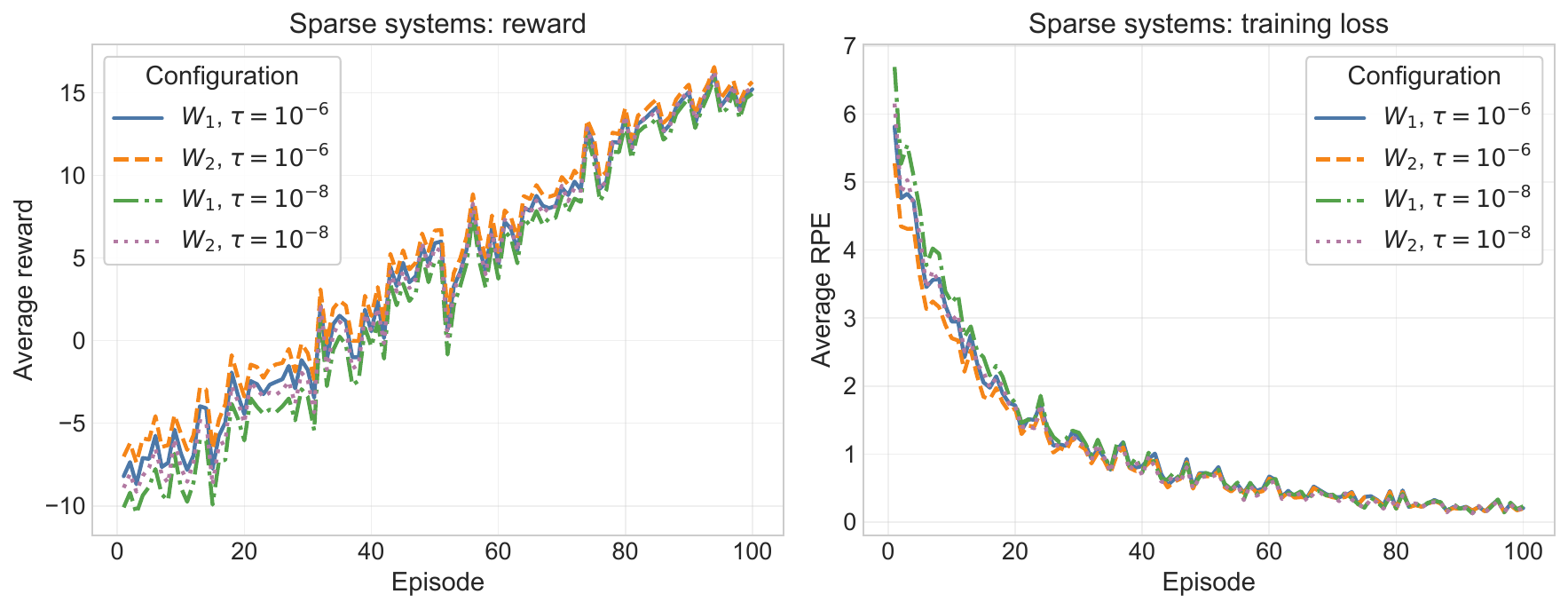}
    \caption{Episode rewards and training loss/RPE for the sparse systems across the four reward-weight/tolerance settings: $(W_1, \tau=10^{-6})$, $(W_2, \tau=10^{-6})$, $(W_1, \tau=10^{-8})$, and $(W_2, \tau=10^{-8})$.}
    \label{fig:sparse_training}
\end{figure}

\begin{figure}[htp!]
    \centering
    \includegraphics[width=0.99\linewidth]{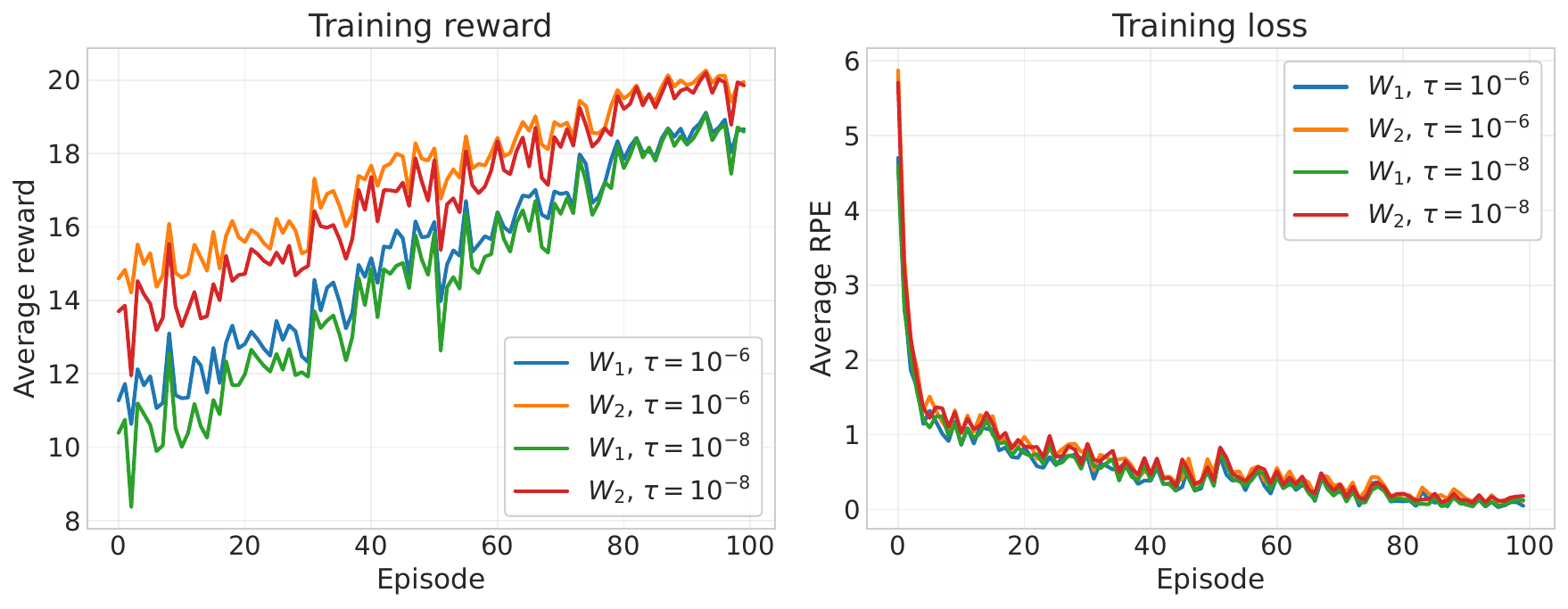}
    \caption{Episode rewards and training loss/RPE for the PDE-generated systems across the four reward-weight/tolerance settings: $(W_1, \tau=10^{-6})$, $(W_2, \tau=10^{-6})$, $(W_1, \tau=10^{-8})$, and $(W_2, \tau=10^{-8})$.}
    \label{fig:pde_training}
\end{figure}

For each condition number range $\mathcal{R}_j$, a threshold is defined:
\begin{equation}\label{eq:sucessful_candidates}
\tau_j = \tau_{\text{base}} \cdot \text{median}(\{ \kappa(A) \mid \mathcal{P} = (A, b) \in \mathcal{R}_j \}),
\end{equation}
where $\tau_{\text{base}} > 0$ is the base tolerance. A system $\mathcal{P} \in \mathcal{R}_j$ is considered successfully solved if:
\begin{equation}
\epsilon_{\text{max}}(\mathcal{P}, \mathbf{a}) < \tau_j.
\end{equation}

To evaluate the proportion of systems that achieve acceptable accuracy, the success rate $\xi_j$ for the range $\mathcal{R}_j$ is:
\begin{equation}\label{eq:sucessful_rate}
\xi_j = \frac{|\{ \mathcal{P} \in \mathcal{R}_j \mid \epsilon_{\text{max}}(\mathcal{P}, \mathbf{a}) < \tau_j \}|}{|\mathcal{R}_j|},
\end{equation}
where $|\mathcal{R}_j|$ is the cardinality $\mathcal{R}_j$, and the threshold $\tau_j$ is scaled by the condition number to account for problem difficulty, aligning with the error bounds in \cite{10.1145/1141885.1141894}.

To comprehensively assess the performance of our approach and study the hyperparameters,  we choose the precision types of BF16, TF32, FP32, and FP64 for exploration and train our models with two weight settings $W_1$ and $W_2$ for \eqref{eq:reward}; $W_1$ indicates $w_1=1, w_2=0.1$, and $W_2$ indicates $w_1=w_2=1.0$. \rev{The accuracy scale is set to $C_1=1$ in \eqref{eq:reward_accuracy}. The dense experiments use $\lambda_{\text{iter}}=1$, while the sparse and PDE-generated experiments use $\lambda_{\text{iter}}=0.5$ with the same logarithmic GMRES-iteration penalty. Compared to $W_1$, $W_2$ increases the weight assigned to the precision-efficiency term and is therefore expected to encourage more aggressive lower-precision selections. Since these experiments were executed as CPU jobs, the BF16 and TF32 results should be interpreted as software-level standard-format precision pathways used to study numerical behavior and learned precision selection, rather than as measurements of accelerator-specific BF16/TF32 throughput. All precision types used in the experiments are supported by the PyTorch library \citep{3454287.3455008} through standard-format pathways. Customized software-defined precision simulation can also be supported by \texttt{pychop}-backed rounding utilities \citep{carson2025}.}

The following simulations evaluate performance using the success rate $\xi$, the normwise relative forward error ($\texttt{ferr}$), the normwise relative backward error ($\texttt{nbe}$), and the iteration count, with comparisons against an FP64 baseline.  All the {\emph{Avg. ferr}}, {\emph{Avg. nbe}}, {\emph{Avg iter.}}, and {\emph{Avg. GMRES iter.}} values in the tables are separately calculated from the average value of $\texttt{ferr}$, $\texttt{nbe}$, converged iterations for iterative refinement, and total GMRES iterations used for the inner solve from all samples, respectively. \xy{The reward and the average Reward Prediction Error (RPE) per episode during training are shown in Figures~\ref{fig:dense_training} and~\ref{fig:sparse_training} and \ref{fig:pde_training}, showing no evidence of overfitting.}

\begin{table*}[h!]
\centering\scriptsize
\setlength{\tabcolsep}{7.5pt}
\renewcommand{\arraystretch}{1.15}
\caption{Average Performance Metrics Across Condition Ranges for Dense Systems.}
\label{tab:pm_ws}
\begin{tabular}{l c c
                S[table-format=1.2e-2]
                S[table-format=1.2e-2]
                S[table-format=1.2]
                S[table-format=1.2]}
\toprule
\textbf{Method} & \textbf{Condition Range} & {$\boldsymbol{\xi}$}
& {\textbf{Avg. ferr}} & {\textbf{Avg. nbe}} & {\textbf{Avg iter.}} & {\textbf{Avg. GMRES iter.}}\\
\midrule

\multicolumn{7}{c}{\vspace{1pt}\textbf{$\tau = 10^{-6}$}}\\
\midrule
\multirow{3}{*}{RL($W_1$)}
 & Low ($10^0$--$10^3$)    & \pct{100} & 1.19e-14 & 7.90e-17 & 2.35 & 2.70 \\
 & Medium ($10^3$--$10^6$) & \pct{100} & 1.57e-11 & 2.82e-15 & 2.26 & 2.20 \\
 & High ($10^6$--$10^9$)   & \pct{100} & 1.89e-09 & 1.63e-14 & 2.21 & 2.54 \\
\midrule
\multirow{3}{*}{RL($W_2$)}
 & Low ($10^0$--$10^3$)    & \pct{100} & 2.48e-07 & 2.19e-08 & 2.30 & 8.00 \\
 & Medium ($10^3$--$10^6$) & \pct{100} & 4.22e-07 & 4.95e-11 & 2.29 & 3.40 \\
 & High ($10^6$--$10^9$)   & \pct{100} & 1.89e-09 & 1.63e-14 & 2.21 & 2.54 \\
\midrule[0.55pt]
\rowcolor{gray!9}
\multicolumn{7}{c}{\textbf{FP64 Baseline ($\tau = 10^{-6}$)}}\\
\rowcolor{gray!9}
\multirow{3}{*}{FP64}
 & Low ($10^0$--$10^3$)    & {--} & 1.20e-14 & 7.96e-17 & 2.00 & 2.00 \\
\rowcolor{gray!9}
 & Medium ($10^3$--$10^6$) & {--} & 1.46e-11 & 8.13e-17 & 2.00 & 2.00 \\
\rowcolor{gray!9}
 & High ($10^6$--$10^9$)   & {--} & 1.92e-09 & 8.09e-17 & 2.00 & 2.00 \\
\midrule

\multicolumn{7}{c}{\vspace{1pt}\textbf{$\tau = 10^{-8}$}}\\
\midrule
\multirow{3}{*}{RL($W_1$)}
 & Low ($10^0$--$10^3$)    & \pct{100} & 1.18e-14 & 7.40e-17 & 2.35 & 4.49 \\
 & Medium ($10^3$--$10^6$) & \pct{100} & 2.95e-11 & 9.60e-13 & 2.14 & 4.77 \\
 & High ($10^6$--$10^9$)   & \pct{100} & 2.02e-09 & 7.93e-17 & 2.11 & 3.79 \\
\midrule
\multirow{3}{*}{RL($W_2$)}
 & Low ($10^0$--$10^3$)    & \pct{89.2} & 2.49e-07 & 2.61e-08 & 2.22 & 9.54 \\
 & Medium ($10^3$--$10^6$) & \pct{100}  & 3.49e-06 & 6.44e-11 & 2.14 & 4.71 \\
 & High ($10^6$--$10^9$)   & \pct{100}  & 3.91e-06 & 4.28e-11 & 2.11 & 3.57 \\
\midrule[0.55pt]
\rowcolor{gray!9}
\multicolumn{7}{c}{\textbf{FP64 Baseline ($\tau = 10^{-8}$)}}\\
\rowcolor{gray!9}
\multirow{3}{*}{FP64}
 & Low ($10^0$--$10^3$)    & {--} & 1.20e-14 & 7.96e-17 & 2.00 & 2.00 \\
\rowcolor{gray!9}
 & Medium ($10^3$--$10^6$) & {--} & 1.46e-11 & 8.13e-17 & 2.00 & 2.00 \\
\rowcolor{gray!9}
 & High ($10^6$--$10^9$)   & {--} & 1.92e-09 & 8.14e-17 & 2.00 & 2.93 \\
\bottomrule
\end{tabular}
\end{table*}

\begin{figure*}[ht!]
    \centering
    \subfigure[RL ($W_1$, $\tau = 10^{-6}$)]{\includegraphics[width=0.42\linewidth]{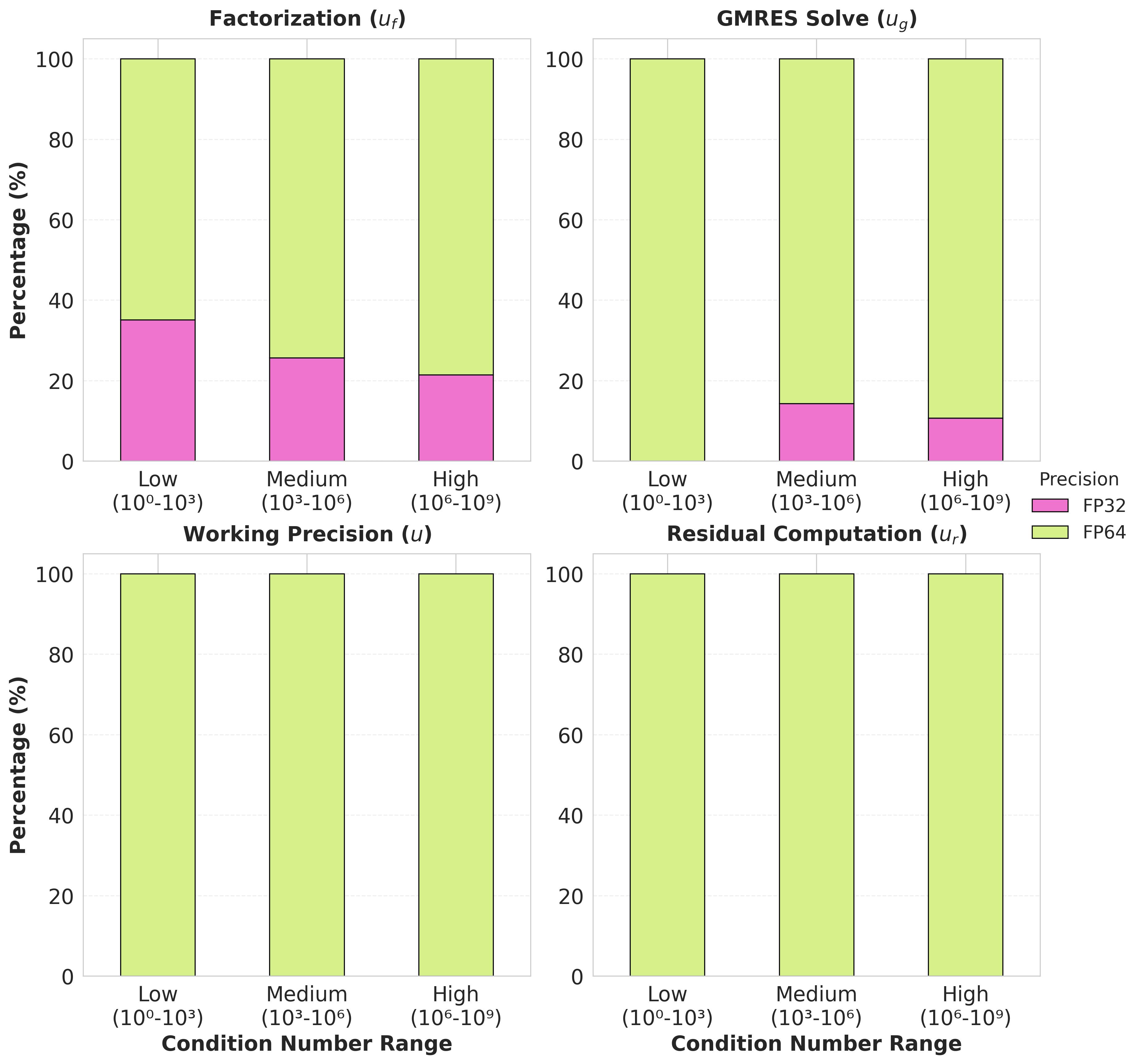}}
    \subfigure[RL ($W_2$, $\tau = 10^{-6}$)]{\includegraphics[width=0.42\linewidth]{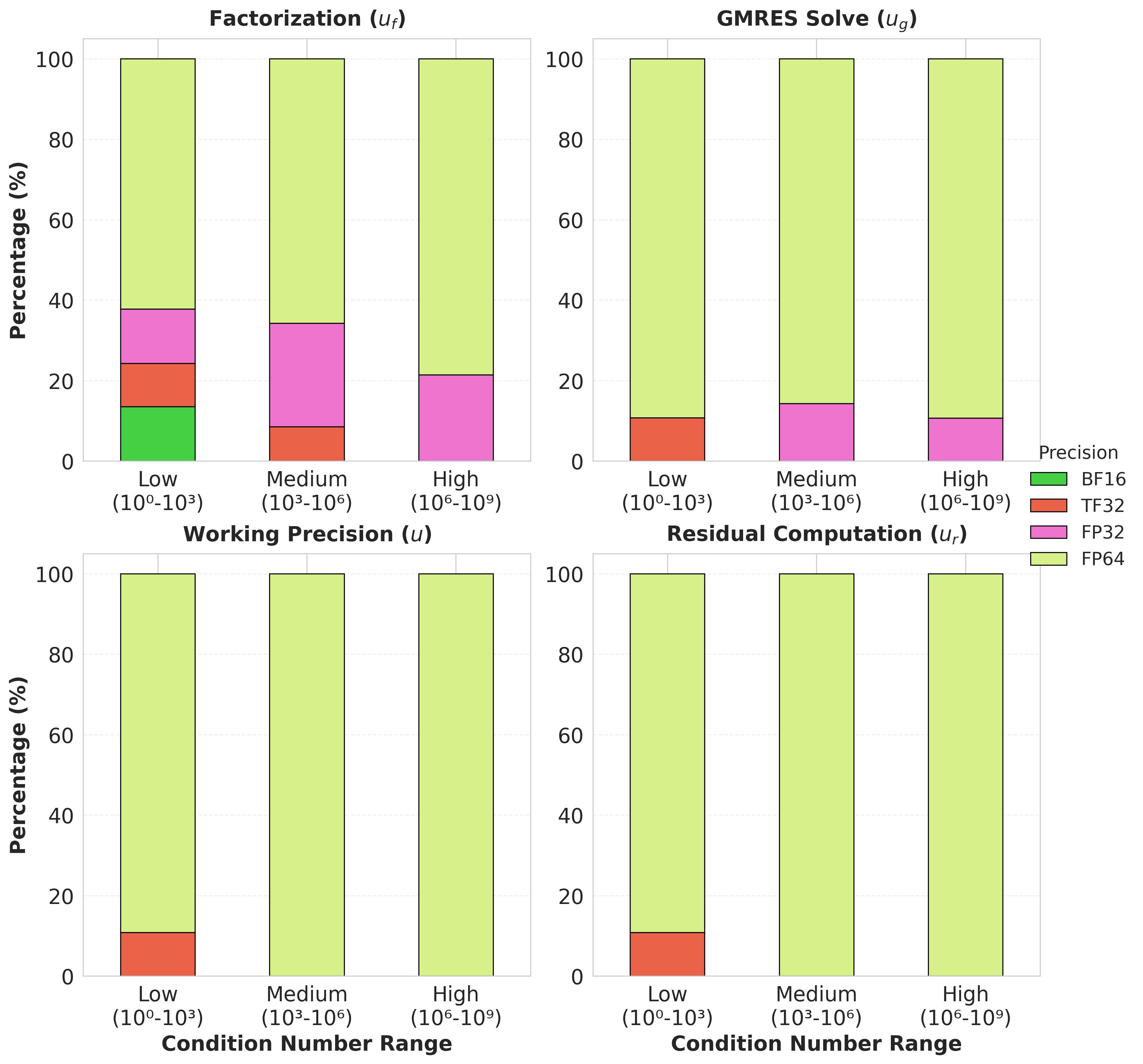}}
    \subfigure[RL ($W_1$, $\tau = 10^{-8}$)]{\includegraphics[width=0.42\linewidth]{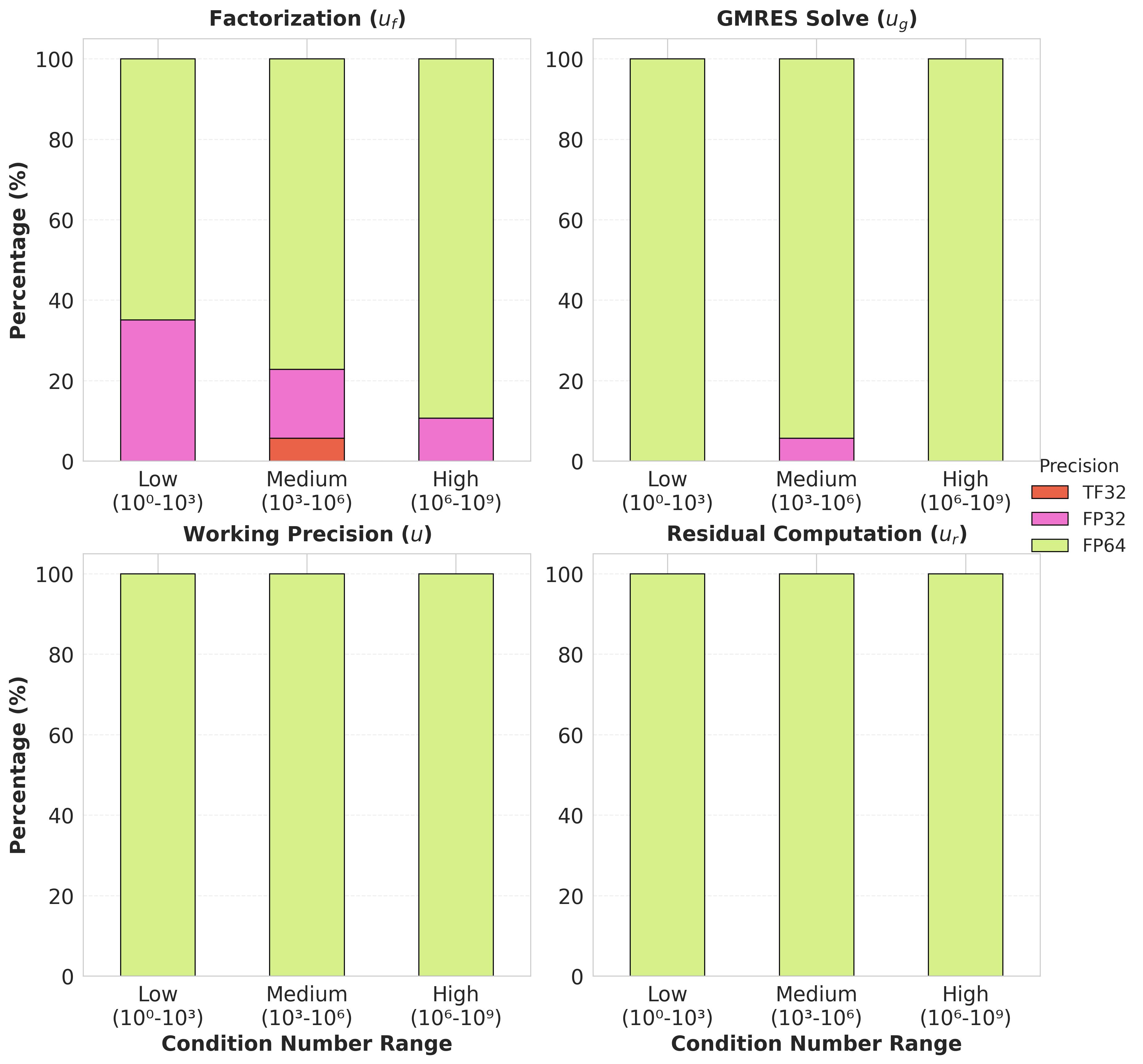}}
    \subfigure[RL ($W_2$, $\tau = 10^{-8}$)]{\includegraphics[width=0.42\linewidth]{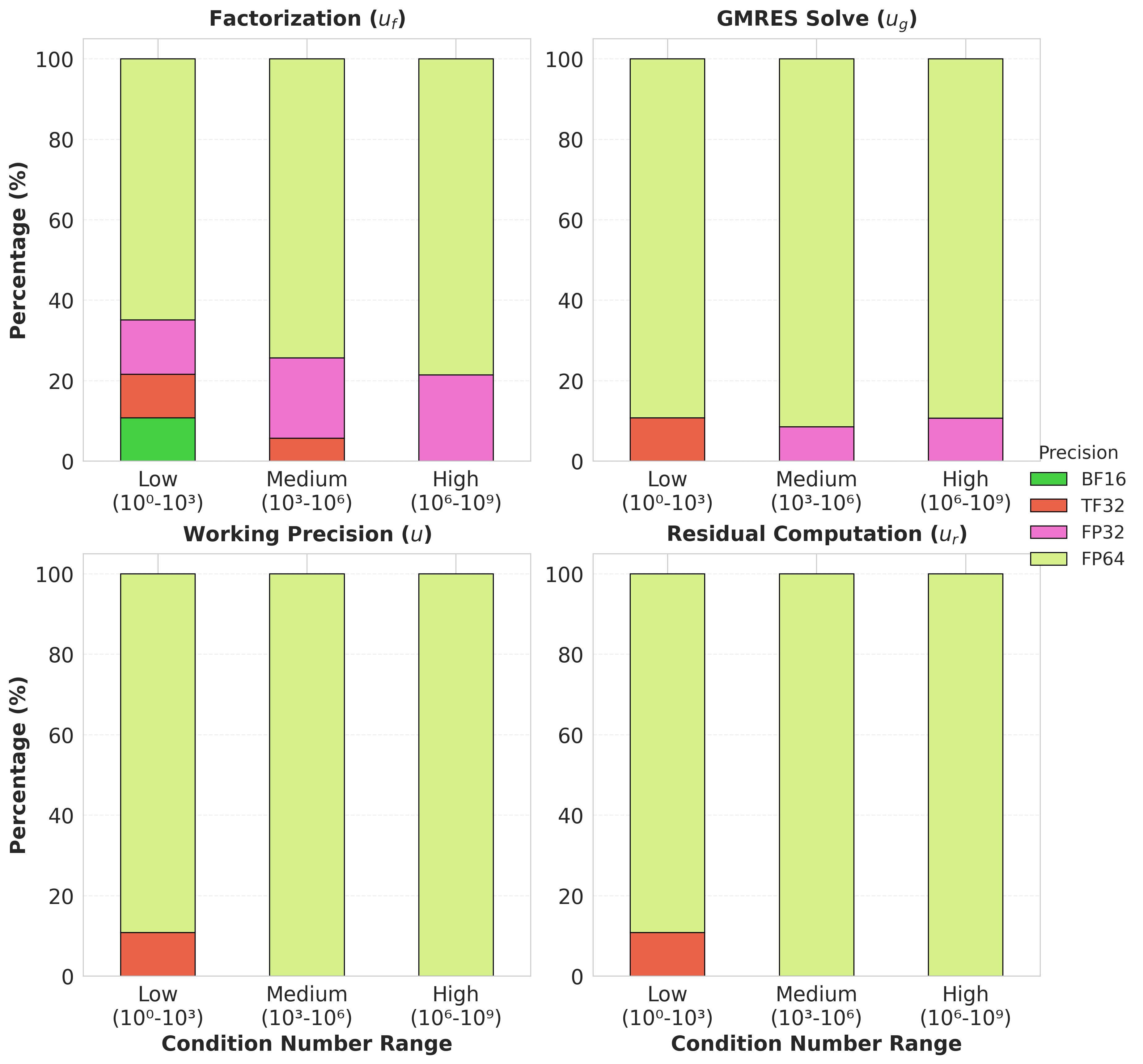}}
    \caption{Average floating-point formats selected across low, medium, and high condition-number ranges for the dense test set. Each panel reports one reward-weight/tolerance setting: $W_1=(w_1,w_2)=(1,0.1)$ or $W_2=(1,1)$, with $\tau \in \{10^{-6},10^{-8}\}$.}
    \label{fig:prec_usage}
\end{figure*}

\subsection{Simulations on Dense Systems}
We generated linear systems with varying condition numbers and matrix sizes for training and testing sets, following MATLAB's gallery('randsvd') function with mode=2. These datasets are created using a singular value decomposition approach. The synthetic sets test solvers with mixed-precision configurations determined by RL. This allows us to verify the reinforcement learning agent's robustness in selecting precision strategies for systems that exhibit variability. For a matrix $A \in \mathbb{R}^{n \times n}$, we construct:
\begin{equation*}
A = U \Sigma V^T,
\end{equation*}
where $U, V \in \mathbb{R}^{n \times n}$ are orthogonal matrices obtained via QR decomposition of random matrices with entries drawn from a standard normal distribution, and $\Sigma = \operatorname{diag}(\sigma_1, \sigma_2, \ldots, \sigma_n)$ is a diagonal matrix with the singular values.

The singular values are:
\begin{equation}
\sigma_k =
\begin{cases}
\sigma_{\text{max}}, & k = 1, 2, \ldots, n-1, \\
\sigma_{\text{max}} / \kappa, & k = n,
\end{cases}
\end{equation}
where $\sigma_{\text{max}} > 0$ is the largest singular value and $\kappa \geq 1$ is the desired condition number, ensuring $\kappa(A) = \sigma_1 / \sigma_n = \kappa$. The singular values are sorted in descending order ($\sigma_1 = \sigma_2 = \cdots = \sigma_{n-1} = \sigma_{\text{max}} \geq \sigma_n = \sigma_{\text{max}} / \kappa$), resulting in a dense matrix $A$ with the specified condition number $\kappa(A) = \kappa$. The ground-truth solution $x \in \mathbb{R}^n$ is generated with entries independently sampled from a standard normal distribution, and the right-hand side vector $b = A x$.

\tablename~\ref{tab:pm_ws} reports error metrics and iteration counts for the linear solve, organized by condition number ranges: low ($10^0$ to $10^3$), medium ($10^3$ to $10^6$), and high ($10^6$ to $10^9$). \figurename~\ref{fig:prec_usage} details the precision types selected within finer condition number intervals.

\begin{figure*}[ht!]
    \centering
    \makebox[\textwidth][c]{%
        \subfigure[$W_2$, $\tau=10^{-6}$]{%
            \includegraphics[width=0.45\textwidth,keepaspectratio]
            {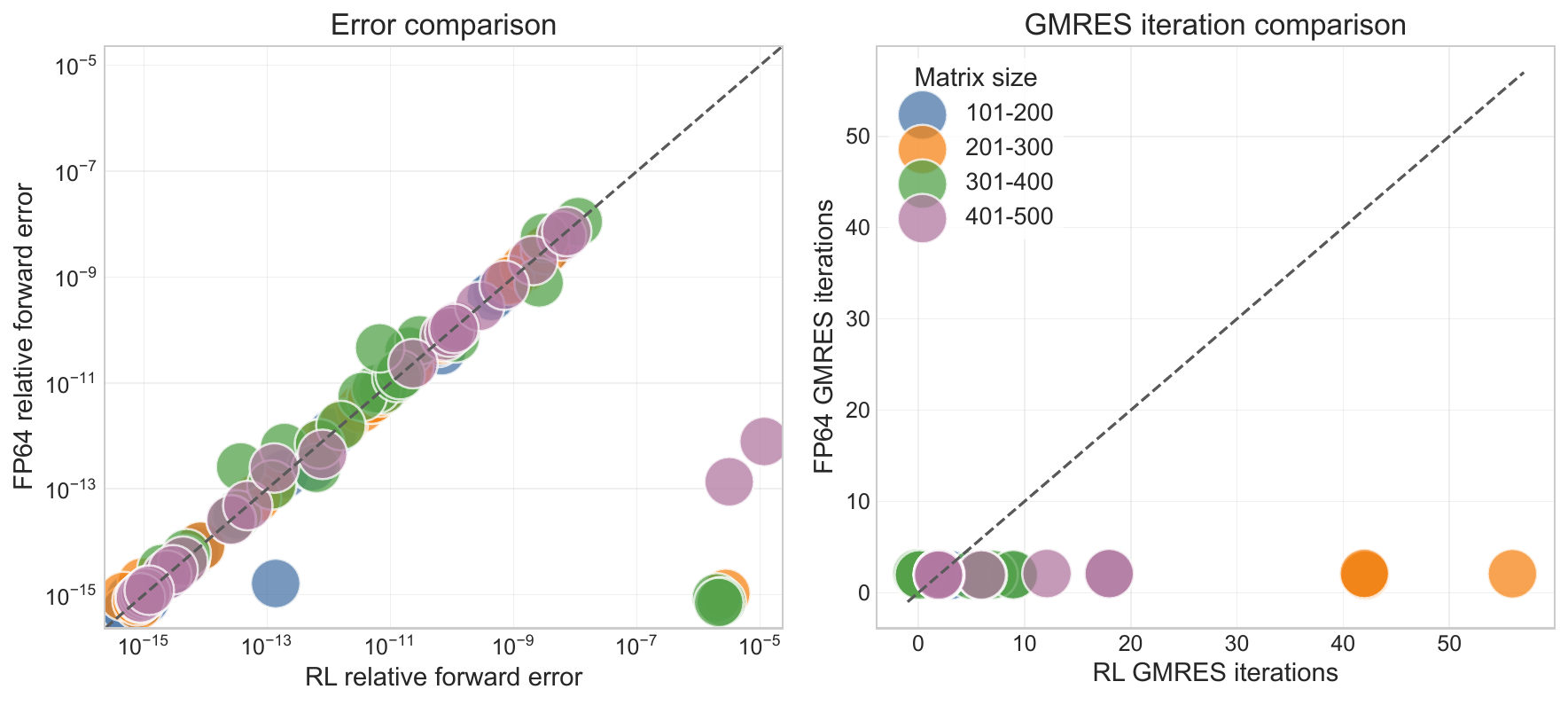}}%
        \hspace{25pt}%
        \subfigure[$W_2$, $\tau=10^{-8}$]{%
            \includegraphics[width=0.45\textwidth,keepaspectratio]
            {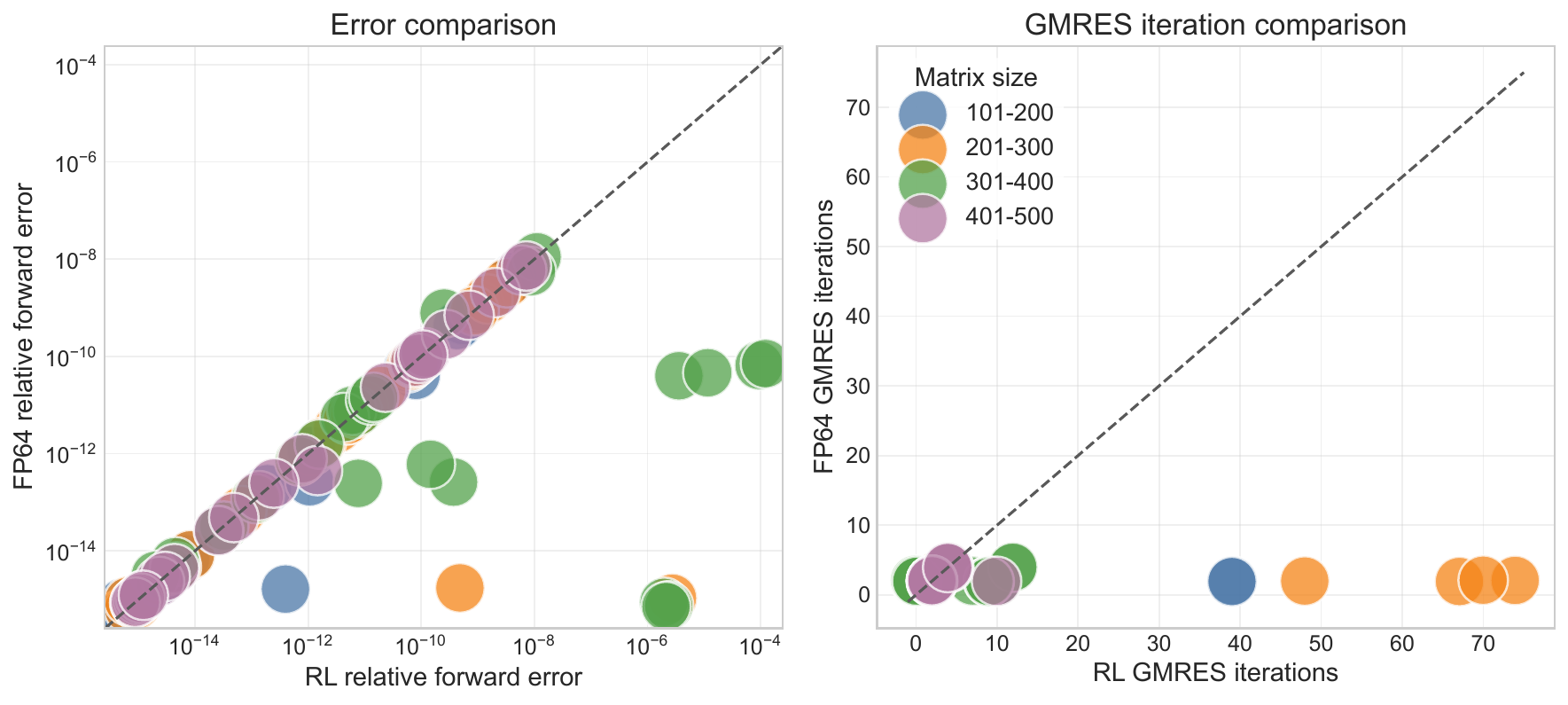}}%
    }
    \caption{Comparison of RL and FP64 in terms of errors and iteration counts.
    Marker colors denote different matrix-size groups.}
    \label{fig:err_iters}
\end{figure*}

Regarding convergence, the conservative $W_1$ policy achieves universal convergence with a success rate $\xi = 100\%$ across all settings. In comparison, the aggressive $W_2$ policy maintains $100\%$ success at $\tau = 10^{-6}$ but shows reduced robustness at the stricter tolerance $\tau = 10^{-8}$, with $\xi$ decreasing to $89.2\%$ in the low-$\kappa$ tests (RL fails in a few test samples). This is likely because the RL had insufficient exposure to training data in this specific range.

With respect to accuracy, $W_1$ closely matches the FP64 baseline, producing forward errors of the same order of magnitude and comparable backward errors. Precision lower than FP32 is never used in $W_1$. By contrast, $W_2$ exhibits a clear trade-off: BF16 and TF32 start to be used within low-to-medium $\kappa$ ranges, which is consistent with the fact that smaller condition numbers theoretically permit lower-precision computations with decreased error amplification. The $W_2$ policy permits the use of low precisions with substantially higher errors (e.g., \texttt{ferr} $\approx 10^{-7}$) to optimize the reward function, as long as the convergence quality is guaranteed.

This behavioral distinction also shows up in computational effort. The average number of outer iterations remains stable ($\approx 2.0$ to $2.35$) for both policies. However, the inner solver cost varies. The $W_2$ policy requires slightly more inner GMRES iterations (e.g., $8.00$ compared to $2.70$ for $W_1$ in low-$\kappa$ bins). This increase is needed to offset the growing rounding errors introduced by lower-precision formats.

The visualization of $\texttt{ferr}$ and total GMRES iterations used for each test sample in $W_2$ is shown in \figurename~\ref{fig:err_iters}. The average performance degradation in \tablename~\ref{tab:pm_ws} for the $W_2$ policy stems from a few instances that deviate from the identity line due to its aggressive strategy, favoring reduced precision.

The precision selection patterns in \figurename~\ref{fig:prec_usage} clarify these performance outcomes. In the low-$\kappa$ range ($10^0$ to $10^3$), $W_1$ employs a conservative combination of FP64 and FP32, entirely avoiding lower precisions. In contrast, $W_2$ leverages BF16 and TF32 (with usage frequencies of approximately $0.33$ and $0.76$, respectively), reflecting a focus on cost-efficiency. As the condition number increases, both strategies converge; beyond $10^6$, both policies adopt an FP64-dominant strategy and avoid the use of lower precisions.

Although our RL framework for precision selection tolerates a loss of accuracy, the primary finding is that the RL agent develops an intelligent condition-dependent precision-adaptation strategy. It utilizes lower-precision formats (BF16/TF32) to achieve computational efficiency in well-conditioned systems, particularly under the $W_2$ policy, but transitions decisively to high-precision FP64 as the matrix becomes ill-conditioned. This behavior demonstrates that the RL agent internalizes theoretical constraints on backward error growth and dynamically prioritizes numerical stability over computational speed when problem difficulty increases.

\subsection{Simulations on Sparse Systems}
In this simulation, we emulated sparse linear systems following the experiment described in \cite{hausner2024neural}. A matrix $A_0 \in \mathbb{R}^{n \times n}$ is constructed with $\text{nnz}(A_0) = \lfloor \lambda_s n^2 \rfloor$ non-zero entries, where $\lambda_s > 0$ is the sparsity parameter. The entries are sampled from a standard normal distribution:
\begin{equation*}
a_{ij} \sim \mathcal{N}(0, 1), \quad (i, j) \in \{(r_k, c_k)\}_{k=1}^{\text{nnz}(A_0)},
\end{equation*}
where $r_k, c_k \in \{1, \ldots, n\}$ are randomly chosen indices. The matrix is symmetrized and made positive definite:
\begin{equation*}
A = A_0 A_0^T + \beta I,
\end{equation*}
where $\beta > 0$ is a diagonal shift and $I$ is the identity matrix. The same as the tests for dense linear systems, the ground-truth solution $x \in \mathbb{R}^n$ is generated with entries independently sampled from a standard normal distribution, and the right-hand side vector $b = A x$.   For all problems, matrices have non-zero diagonal entries ($a_{ii} \neq 0$) and are non-singular.

\begin{table*}[ht!]
\centering\footnotesize
\caption{Train/Test Metrics Summary}\label{tab:sparse_stats}
\begin{tabular}{lcc}
\toprule
\textbf{Metric} & \textbf{Train (min -- max)} & \textbf{Test (min -- max)} \\
\midrule
Condition number & $9.926\times10^{7}$ -- $1.559\times10^{10}$ & $1.016\times10^{8}$ -- $7.863\times10^{9}$ \\
Sparsity          & 1.80\% -- 5.10\% & 1.80\% -- 5.00\% \\
Matrix size      & 103 -- 500 & 100 -- 498 \\
\bottomrule
\end{tabular}
\end{table*}

Setting $\lambda_s=0.01$, the random sparse linear systems in our training set and testing set following the above generation routines are described in \tablename~\ref{tab:sparse_stats}. Similar to the above (under the same evaluation metrics, two tolerance levels, and two reward weight policies), \tablename~\ref{tab:pm_ws_sparse} showcases the average performance of the RL-guided mixed-precision iterative refinement solver on sparse linear systems. \tablename~\ref{tab:prec_usage_sparse} details the average selection frequency of four floating-point formats across the four precision-controlled stages, with each row summing to 4.


\begin{table*}[h!]
\centering\scriptsize
\setlength{\tabcolsep}{7.5pt}
\renewcommand{\arraystretch}{1.15}

\caption{Average Performance Metrics for Sparse Systems.}
\label{tab:pm_ws_sparse}

\begin{tabular}{l
                S[table-format=3.2]
                S[table-format=1.2e-1]
                S[table-format=1.2e-1]
                S[table-format=1.2]
                S[table-format=1.2]}
\toprule
\textbf{Method}
& {$\boldsymbol{\xi(\%)}$}
& {\textbf{Avg. ferr}}
& {\textbf{Avg. nbe}}
& {\textbf{Avg Iter.}}
& {\textbf{Avg. GMRES iter.}}\\
\midrule

\multicolumn{6}{c}{\vspace{1pt}\textbf{$\tau = 10^{-6}$}}\\
\midrule
RL ($W_1$)       & 100.00 & 4.91e-09 & 5.26e-17 & 2.01 & 2.22 \\
RL ($W_2$)       & 100.00 & 4.91e-09 & 5.26e-17 & 2.01 & 2.22 \\
\rowcolor{gray!9}
FP64 Baseline    & {--}   & 4.90e-09 & 5.28e-17 & 2.00 & 2.00 \\
\midrule[0.55pt]

\multicolumn{6}{c}{\vspace{1pt}\textbf{$\tau = 10^{-8}$}}\\
\midrule
RL ($W_1$)       & 100.00 & 4.90e-09 & 5.28e-17 & 2.00 & 2.10 \\
RL ($W_2$)       & 100.00 & 4.91e-09 & 5.28e-17 & 2.01 & 2.36 \\
\rowcolor{gray!9}
FP64 Baseline    & {--}   & 4.90e-09 & 5.28e-17 & 2.00 & 2.10 \\
\bottomrule
\end{tabular}
\end{table*}

\begin{table}[h!]
\centering
\caption{Average Floating-point Precision Usage Per Solve for Sparse Systems.}
\label{tab:prec_usage_sparse}
\scriptsize
\setlength{\tabcolsep}{7pt}
\renewcommand{\arraystretch}{1.15}

\begin{tabular}{l
                S[table-format=1.2]
                S[table-format=1.2]
                S[table-format=1.2]
                S[table-format=1.2]}
\toprule
\textbf{Weight Setting} & \textbf{BF16} & \textbf{TF32} & \textbf{FP32} & \textbf{FP64} \\
\midrule

\multicolumn{5}{c}{\textbf{$\tau = 10^{-6}$}}\\
\midrule
{RL ($W_1$)} & 0.00 & 0.00 & 0.01 & 3.99 \\
{RL ($W_2$)} & 0.00 & 0.00 & 0.01 & 3.99 \\
\midrule

\multicolumn{5}{c}{\textbf{$\tau = 10^{-8}$}}\\
\midrule
{RL ($W_1$)} & 0.00 & 0.00 & 0.00 & 4.00 \\
{RL ($W_2$)} & 0.00 & 0.00 & 0.01 & 3.99 \\
\bottomrule
\end{tabular}
\end{table}

The evaluation of sparse linear systems presents a distinct contrast to the dense case, primarily driven by the inherent numerical difficulty posed by the very ill-conditioned dataset. As depicted in \tablename~\ref{tab:sparse_stats}, the test matrices exhibit high condition numbers between $1.02{\times}10^{8}$ and $7.86{\times}10^{9}$, which places linear solve in a range where numerical stability is precarious, leaving limited room for precision reduction. This test effectively demonstrates whether the RL agent can effectively avoid numerical instabilities in its precision selection.

\tablename~\ref{tab:pm_ws_sparse} summarizes the performance metrics. Under these harsh conditions, the RL agent exhibits a highly conservative behavior. Regardless of the reward weight setting ($W_1$ or $W_2$) or the tolerance level ($\tau$), the agent achieves a $100\%$ success rate with error metrics (\texttt{ferr} and \texttt{nbe}) that are close to the FP64 baseline.
Unlike the dense case, where $W_2$ triggered an increase in GMRES iterations to trade off lower precision, here the average iteration counts (both outer and GMRES) for RL policies align closely with the baseline. This indicates that the agent has learned that any reduction in precision, even for the sake of potential reward, would likely lead to stagnation or divergence.

This conservative approach is further quantified in Table~\ref{tab:prec_usage_sparse}. The precision distribution indicates a consistent reliance on double precision, with FP64 usage averaging approximately $3.99$ to $4.00$ steps per solve. Even under the aggressive weighting scheme $W_2$, which strongly incentivizes low-precision usage, the agent entirely avoids selecting BF16 and TF32.
This convergence of $W_1$ and $W_2$ strategies is a significant finding: it demonstrates that the RL agent correctly identifies the "survival boundary" of the iterative method. When faced with uniformly ill-conditioned sparse systems, the policy effectively overrides the efficiency-driven incentives of $W_2$ and prioritizes the fundamental necessity of numerical convergence. Thus, the framework proves robust, defaulting to a high-precision fallback when the problem's spectral properties render low-precision formats numerically unsafe.

\subsection{Simulations on PDE matrices}
\xy{To further test whether the learned precision policy transfers beyond unstructured random matrices, we add a PDE-generated sparse-matrix benchmark following the matrix families used in \cite{carson2026GNNs}. Let $\Omega=(0,1)^2$ and let $G_h=\{(ih,jh):1\leq i,j\leq m\}$ be the interior grid with mesh width $h=(m+1)^{-1}$ and $n=m^2$ unknowns. After lexicographic vectorization, we generate matrices from four finite-difference discretizations:
\begin{align*}
-\Delta u &= f, \\
-\nabla\cdot(K_\epsilon \nabla u) &= f, \qquad K_\epsilon=\operatorname{diag}(\epsilon,1), \\
-\nabla\cdot(\kappa_h(x,y)\nabla u) &= f, \\
-\epsilon \Delta u + \boldsymbol{\beta}\cdot \nabla u &= f.
\end{align*}
The first operator is the standard five-point Poisson stencil. The second is an anisotropic diffusion operator with $\epsilon$ sampled logarithmically from $[10^{-8},10^{-3}]$. The third is a high-contrast diffusion operator in which the cell coefficients $\kappa_h$ are sampled logarithmically between $1$ and a contrast parameter sampled up to $10^{13}$; interface coefficients are formed using harmonic averaging and homogeneous Dirichlet boundary contributions are included on the diagonal. The fourth is a nonsymmetric convection-diffusion operator with $\epsilon\in[10^{-3},1]$ and $\boldsymbol{\beta}=(\beta_x,\beta_y)$ sampled from $[-10,10]^2$, discretized with a first-order upwind treatment of the advective terms.

For each problem, the target matrix size is sampled from $[100,500]$ and rounded upward to the nearest perfect square $m^2$. The right-hand side is generated by a manufactured smooth solution, rather than by sampling $b$ directly. Specifically, for grid coordinates $(x_i,y_j)\in G_h$, we set
\begin{equation*}
x_{ij}=\sin(\pi x_i)\sin(\pi y_j)+\frac{1}{4}\sin(2\pi x_i)\sin(3\pi y_j),
\end{equation*}
and then define $b=Ax$. This construction preserves a known reference solution for computing \texttt{ferr}, while allowing the matrix family to determine the conditioning, sparsity pattern, symmetry, and nonsymmetry of the solve.

\begin{table*}[ht!]
\centering\scriptsize
\setlength{\tabcolsep}{6pt}
\renewcommand{\arraystretch}{1.15}
\caption{Summary of the PDE-generated matrix dataset. The four PDE families are equally represented in the test set.}
\label{tab:pde_stats}
\begin{tabular}{l c c c c}
\toprule
\textbf{Subset/family} & \textbf{Count} & \textbf{Condition number range} & \textbf{Matrix size range} & \textbf{Sparsity range} \\
\midrule
Train overall & 100 & $2.187{\times}10^{1}$ -- $3.592{\times}10^{11}$ & 100 -- 529 & 0.91\% -- 4.60\% \\
Test overall  & 100 & $2.236{\times}10^{1}$ -- $2.222{\times}10^{12}$ & 121 -- 529 & 0.91\% -- 3.83\% \\
\midrule
Poisson 2D & 25 & $5.770{\times}10^{1}$ -- $2.137{\times}10^{2}$ & 121 -- 484 & 1.00\% -- 3.83\% \\
Anisotropic diffusion & 25 & $5.770{\times}10^{1}$ -- $2.328{\times}10^{2}$ & 121 -- 529 & 0.91\% -- 3.83\% \\
High-contrast diffusion & 25 & $6.022{\times}10^{5}$ -- $2.222{\times}10^{12}$ & 196 -- 529 & 0.91\% -- 2.41\% \\
Convection-diffusion & 25 & $2.236{\times}10^{1}$ -- $1.868{\times}10^{2}$ & 121 -- 529 & 0.91\% -- 3.83\% \\
\bottomrule
\end{tabular}
\end{table*}

\begin{figure}[ht!]
    \centering
    \includegraphics[width=0.8\linewidth]{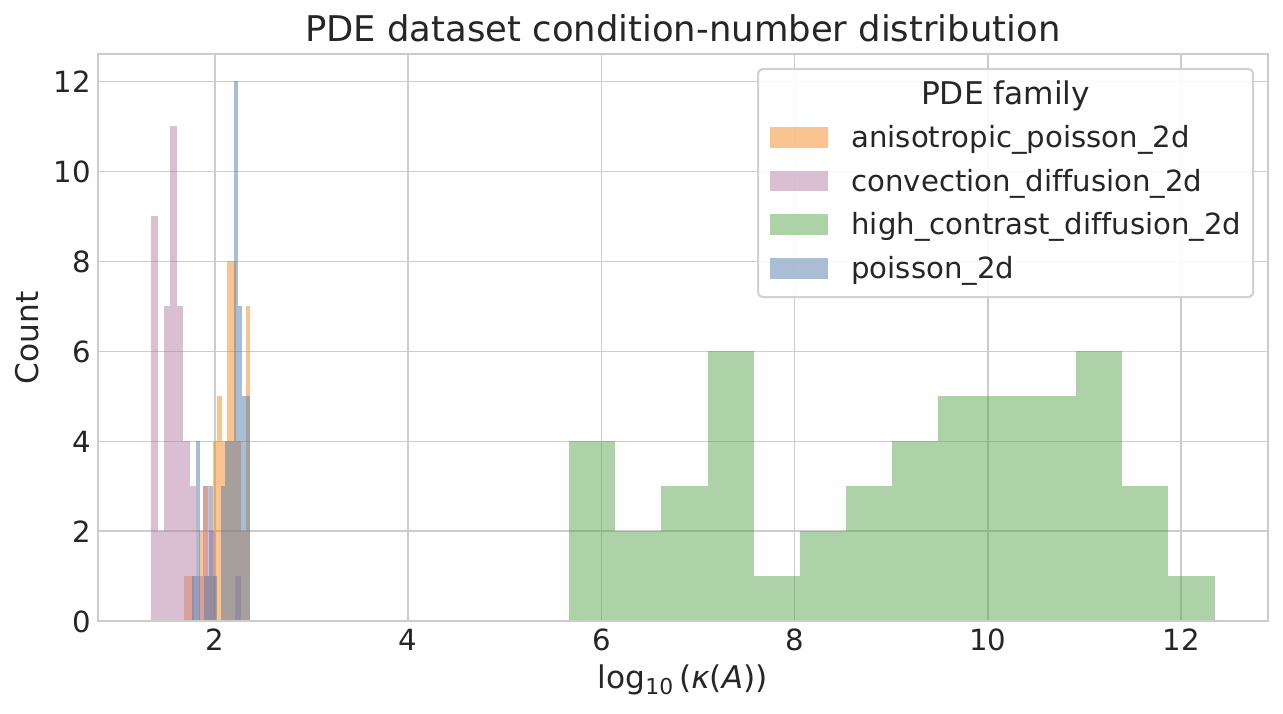}
    \caption{Distribution of condition numbers in the PDE-generated dataset, separated by PDE family. The high-contrast diffusion problems create the ill-conditioned tail, while the Poisson, anisotropic, and convection-diffusion families occupy lower condition-number ranges for the tested grid sizes.}
    \label{fig:pde_cond_dist}
\end{figure}

\tablename~\ref{tab:pde_stats} and \figurename~\ref{fig:pde_cond_dist} show that the PDE dataset is structurally different from the random sparse dataset. Most Poisson, anisotropic, and convection-diffusion matrices are moderately conditioned at these sizes, whereas high-contrast diffusion produces a long ill-conditioned tail extending beyond $10^{12}$. This creates a mixed testbed in which the agent sees both benign PDE operators and PDE operators whose coefficient jumps make low precision risky.
}
\begin{table*}[ht!]
\centering\scriptsize
\setlength{\tabcolsep}{7.5pt}
\renewcommand{\arraystretch}{1.15}
\caption{Average performance metrics for PDE-generated systems. Success rates use the same condition-scaled criterion as in \eqref{eq:sucessful_candidates}--\eqref{eq:sucessful_rate}, with an additional very-high range for $\kappa(A)>10^9$.}
\label{tab:pm_ws_pde}
\begin{tabular}{l
                S[table-format=3.2]
                S[table-format=1.2e-2]
                S[table-format=1.2e-2]
                S[table-format=2.2]
                S[table-format=2.2]}
\toprule
\textbf{Method}
& {$\boldsymbol{\xi(\%)}$}
& {\textbf{Avg. ferr}}
& {\textbf{Avg. nbe}}
& {\textbf{Avg Iter.}}
& {\textbf{Avg. GMRES iter.}}\\
\midrule
\multicolumn{6}{c}{\vspace{1pt}\textbf{$\tau = 10^{-6}$}}\\
\midrule
RL ($W_1$)       & 100.00 & 1.13e-2 & 2.38e-7 & 2.05 & 12.75 \\
RL ($W_2$)       & 100.00 & 1.13e-2 & 2.38e-7 & 2.31 & 13.01 \\
\rowcolor{gray!9}
FP64 Baseline    & {--}   & 3.89e-11 & 6.46e-17 & 2.00 & 2.00 \\
\midrule[0.55pt]
\multicolumn{6}{c}{\vspace{1pt}\textbf{$\tau = 10^{-8}$}}\\
\midrule
RL ($W_1$)       & 100.00 & 4.43e-11 & 8.41e-15 & 2.04 & 5.13 \\
RL ($W_2$)       & 100.00 & 4.43e-11 & 8.41e-15 & 2.04 & 5.13 \\
\rowcolor{gray!9}
FP64 Baseline    & {--}   & 3.89e-11 & 6.46e-17 & 2.00 & 2.00 \\
\bottomrule
\end{tabular}
\end{table*}

\begin{table}[ht!]
\centering
\caption{Average floating-point precision usage per solve for PDE-generated systems.}
\label{tab:prec_usage_pde}
\scriptsize
\setlength{\tabcolsep}{8pt}
\renewcommand{\arraystretch}{1.15}
\begin{tabular}{l
                S[table-format=1.2]
                S[table-format=1.2]
                S[table-format=1.2]
                S[table-format=1.2]}
\toprule
\textbf{Weight setting} & \textbf{BF16} & \textbf{TF32} & \textbf{FP32} & \textbf{FP64} \\
\midrule
\multicolumn{5}{c}{\textbf{$\tau = 10^{-6}$}}\\
\midrule
{RL ($W_1$)} & 0.01 & 0.00 & 0.07 & 3.92 \\
{RL ($W_2$)} & 0.01 & 0.00 & 0.33 & 3.66 \\
\midrule
\multicolumn{5}{c}{\textbf{$\tau = 10^{-8}$}}\\
\midrule
{RL ($W_1$)} & 0.00 & 0.01 & 0.07 & 3.92 \\
{RL ($W_2$)} & 0.00 & 0.01 & 0.07 & 3.92 \\
\bottomrule
\end{tabular}
\end{table}

\begin{figure*}[ht!]
    \centering
    \makebox[\textwidth][c]{%
        \subfigure[$W_2$, $\tau=10^{-6}$]{%
            \includegraphics[width=0.45\textwidth,keepaspectratio]
            {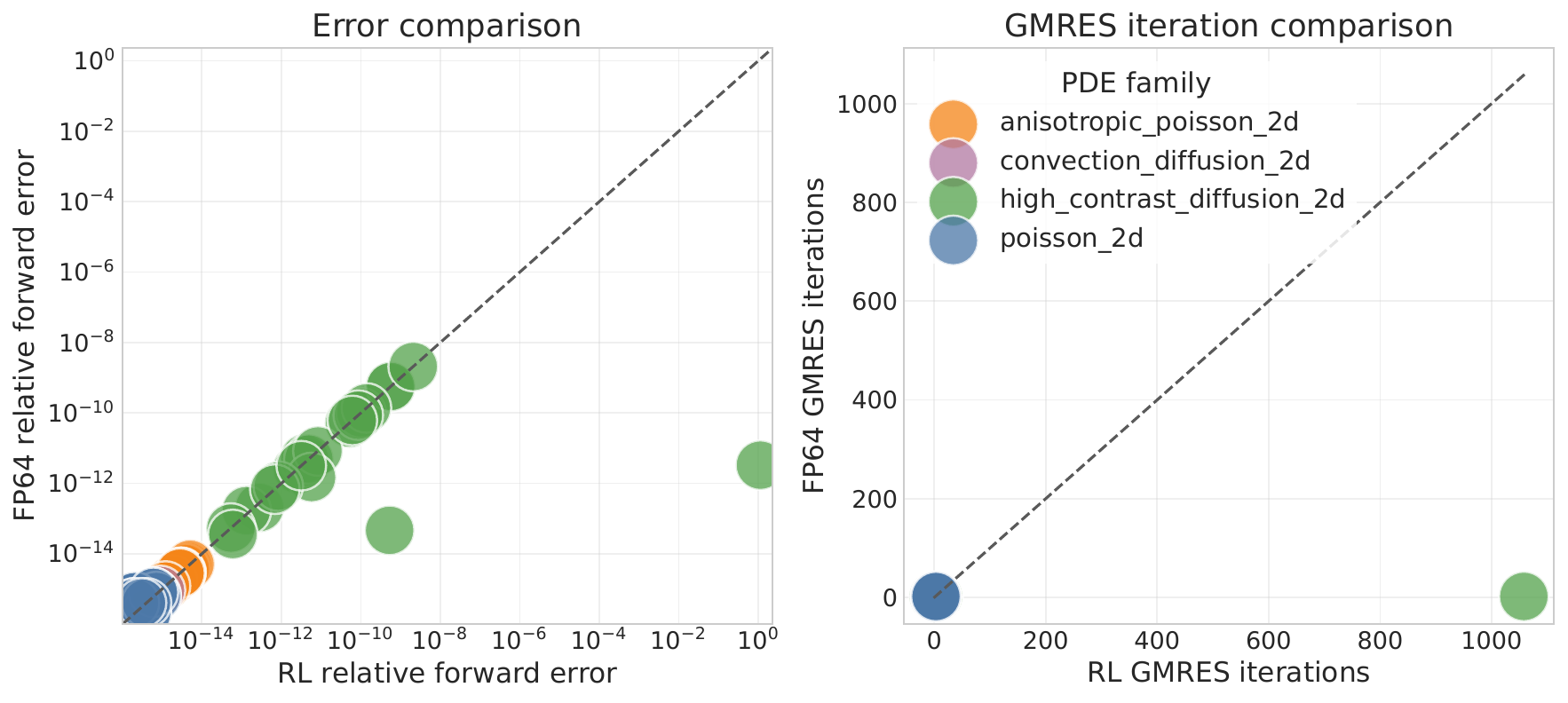}}%
        \hspace{25pt}%
        \subfigure[$W_2$, $\tau=10^{-8}$]{%
            \includegraphics[width=0.45\textwidth,keepaspectratio]
            {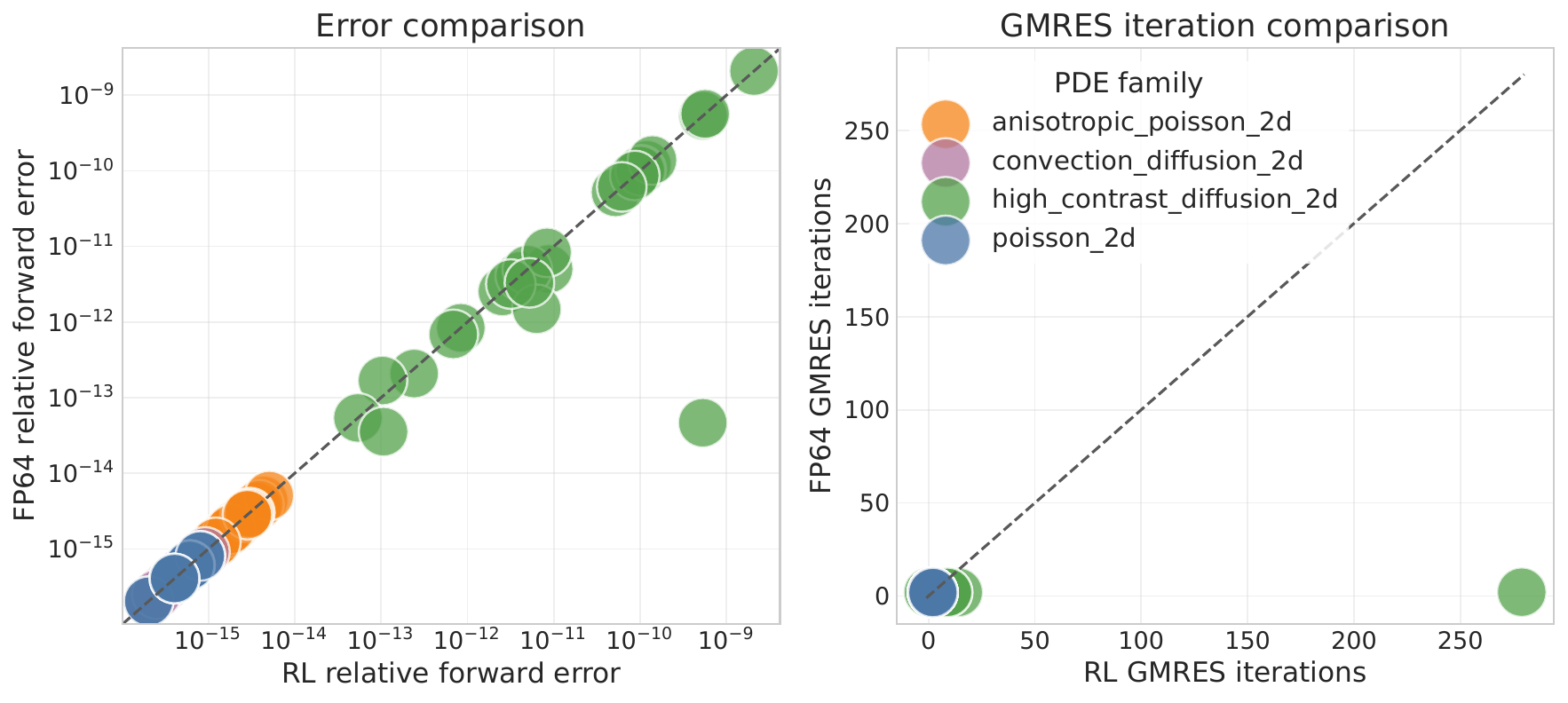}}%
    }
    \caption{Comparison of RL and FP64 on PDE-generated systems under the more aggressive $W_2$ reward setting. Marker colors denote PDE families and marker sizes scale with matrix size.}
    \label{fig:pde_err_iters}
\end{figure*}

\xy{\tablename~\ref{tab:pm_ws_pde} reports the aggregate PDE performance. At $\tau=10^{-6}$, both $W_1$ and $W_2$ achieve $100\%$ success under the condition-scaled criterion, but their average forward errors are larger than the FP64 baseline because the high-contrast diffusion subset contains several difficult coefficient-jump problems. On this subset alone, the average \texttt{ferr} is $4.53{\times}10^{-2}$ for both RL policies at $\tau=10^{-6}$, whereas the Poisson, anisotropic, and convection-diffusion families remain at approximately FP64 accuracy. This is also visible in \figurename~\ref{fig:pde_err_iters}, where the outlying PDE points are associated with the high-contrast family rather than with all sparse PDE matrices.

The stricter tolerance $\tau=10^{-8}$ changes the learned behavior substantially. Both reward settings produce nearly identical precision choices and reach an average \texttt{ferr} of $4.43{\times}10^{-11}$, close to the FP64 baseline value $3.89{\times}10^{-11}$. The main cost is additional inner work: the average total GMRES iteration count increases from $2.00$ for FP64 to $5.13$ for RL, and within the high-contrast family it reaches $14.52$ on average. Thus, the agent compensates for more delicate PDE systems by spending extra Krylov iterations while still satisfying the error criterion.

The precision usage in \tablename~\ref{tab:prec_usage_pde} further confirms that the policy adapts to the PDE difficulty. Under $W_2$ and $\tau=10^{-6}$, the agent replaces part of the FP64 usage by FP32, increasing average FP32 selections from $0.07$ to $0.33$ per solve. However, BF16 and TF32 remain almost unused, and when the tolerance is tightened to $10^{-8}$ the $W_1$ and $W_2$ policies collapse to the same FP64-dominant choice. Compared with the dense experiments, where $W_2$ can safely exploit more reduced precision on low-$\kappa$ problems, the PDE experiments show a more cautious behavior: structured sparsity alone does not determine the precision budget; coefficient contrast and the resulting conditioning are decisive.

Overall, the PDE simulations support the central interpretation of the framework. The contextual bandit does not merely learn a fixed preference for low precision; it learns when reduced precision is numerically admissible. For benign PDE operators it preserves FP64-level accuracy with essentially the same iteration count as the baseline, while for high-contrast diffusion it shifts toward conservative precision choices and, when necessary, pays extra GMRES iterations to maintain stability.}

\begin{table*}[htb!]
\centering\scriptsize
\setlength{\tabcolsep}{7.5pt}
\renewcommand{\arraystretch}{1.15}
\caption{Average performance metrics across condition ranges for dense systems when the iteration penalty $f_{\mathrm{penalty}}$ is removed from the reward. The ``Avg. GMRES iter.'' column again reports the total inner GMRES iterations accumulated over all outer refinement steps, averaged over the corresponding test subset.}
\label{tab:pm_ws_ablation}
\begin{tabular}{l c c
                S[table-format=1.2e-2]
                S[table-format=1.2e-2]
                S[table-format=1.2]
                S[table-format=1.2]}
\toprule
\textbf{Method} & \textbf{Condition Range} & {$\boldsymbol{\xi}$}
& {\textbf{Avg. ferr}} & {\textbf{Avg. nbe}} & {\textbf{Avg iter.}} & {\textbf{Avg. GMRES iter.}}\\
\midrule
\multicolumn{7}{c}{\vspace{1pt}\textbf{$\tau = 10^{-6}$}}\\
\midrule
\multirow{3}{*}{RL($W_1$)}
 & Low ($10^0$--$10^3$) & \pct{100} & 8.86e-08 & 1.45e-10 & 2.65 & 11.05 \\
 & Medium ($10^3$--$10^6$) & \pct{100} & 4.22e-07 & 4.97e-11 & 2.54 & 5.94 \\
 & High ($10^6$--$10^9$) & \pct{100} & 1.94e-09 & 1.63e-14 & 2.32 & 3.25 \\
\midrule
\multirow{3}{*}{RL($W_2$)}
 & Low ($10^0$--$10^3$) & \pct{100} & 3.37e-07 & 2.20e-08 & 2.46 & 12.08 \\
 & Medium ($10^3$--$10^6$) & \pct{100} & 4.22e-07 & 4.97e-11 & 2.54 & 5.94 \\
 & High ($10^6$--$10^9$) & \pct{100} & 1.94e-09 & 1.63e-14 & 2.32 & 3.25 \\
\midrule[0.55pt]
\rowcolor{gray!9}
\multicolumn{7}{c}{\textbf{FP64 Baseline ($\tau = 10^{-6}$)}}\\
\rowcolor{gray!9}
\multirow{3}{*}{FP64}
 & Low ($10^0$--$10^3$) & {--} & 1.20e-14 & 7.96e-17 & 2.00 & 2.00 \\
\rowcolor{gray!9}
 & Medium ($10^3$--$10^6$) & {--} & 1.46e-11 & 8.13e-17 & 2.00 & 2.00 \\
\rowcolor{gray!9}
 & High ($10^6$--$10^9$) & {--} & 1.92e-09 & 8.09e-17 & 2.00 & 2.00 \\
\midrule
\multicolumn{7}{c}{\vspace{1pt}\textbf{$\tau = 10^{-8}$}}\\
\midrule
\multirow{3}{*}{RL($W_1$)}
 & Low ($10^0$--$10^3$) & \pct{100} & 1.34e-11 & 1.62e-12 & 2.38 & 15.97 \\
 & Medium ($10^3$--$10^6$) & \pct{100} & 3.49e-06 & 6.44e-11 & 2.31 & 12.37 \\
 & High ($10^6$--$10^9$) & \pct{100} & 3.91e-06 & 4.28e-11 & 2.11 & 3.57 \\
\midrule
\multirow{3}{*}{RL($W_2$)}
 & Low ($10^0$--$10^3$) & \pct{89.2} & 2.49e-07 & 2.61e-08 & 2.27 & 21.03 \\
 & Medium ($10^3$--$10^6$) & \pct{100} & 3.49e-06 & 6.48e-11 & 2.34 & 12.51 \\
 & High ($10^6$--$10^9$) & \pct{100} & 3.91e-06 & 4.28e-11 & 2.18 & 4.29 \\
\midrule[0.55pt]
\rowcolor{gray!9}
\multicolumn{7}{c}{\textbf{FP64 Baseline ($\tau = 10^{-8}$)}}\\
\rowcolor{gray!9}
\multirow{3}{*}{FP64}
 & Low ($10^0$--$10^3$) & {--} & 1.20e-14 & 7.96e-17 & 2.00 & 2.00 \\
\rowcolor{gray!9}
 & Medium ($10^3$--$10^6$) & {--} & 1.46e-11 & 8.13e-17 & 2.00 & 2.00 \\
\rowcolor{gray!9}
 & High ($10^6$--$10^9$) & {--} & 1.92e-09 & 8.14e-17 & 2.00 & 2.93 \\
\bottomrule
\end{tabular}
\end{table*}

\begin{figure*}[ht!]
    \centering
    \subfigure[RL ($W_1$, $\tau = 10^{-6}$)]{\includegraphics[width=0.43\linewidth]{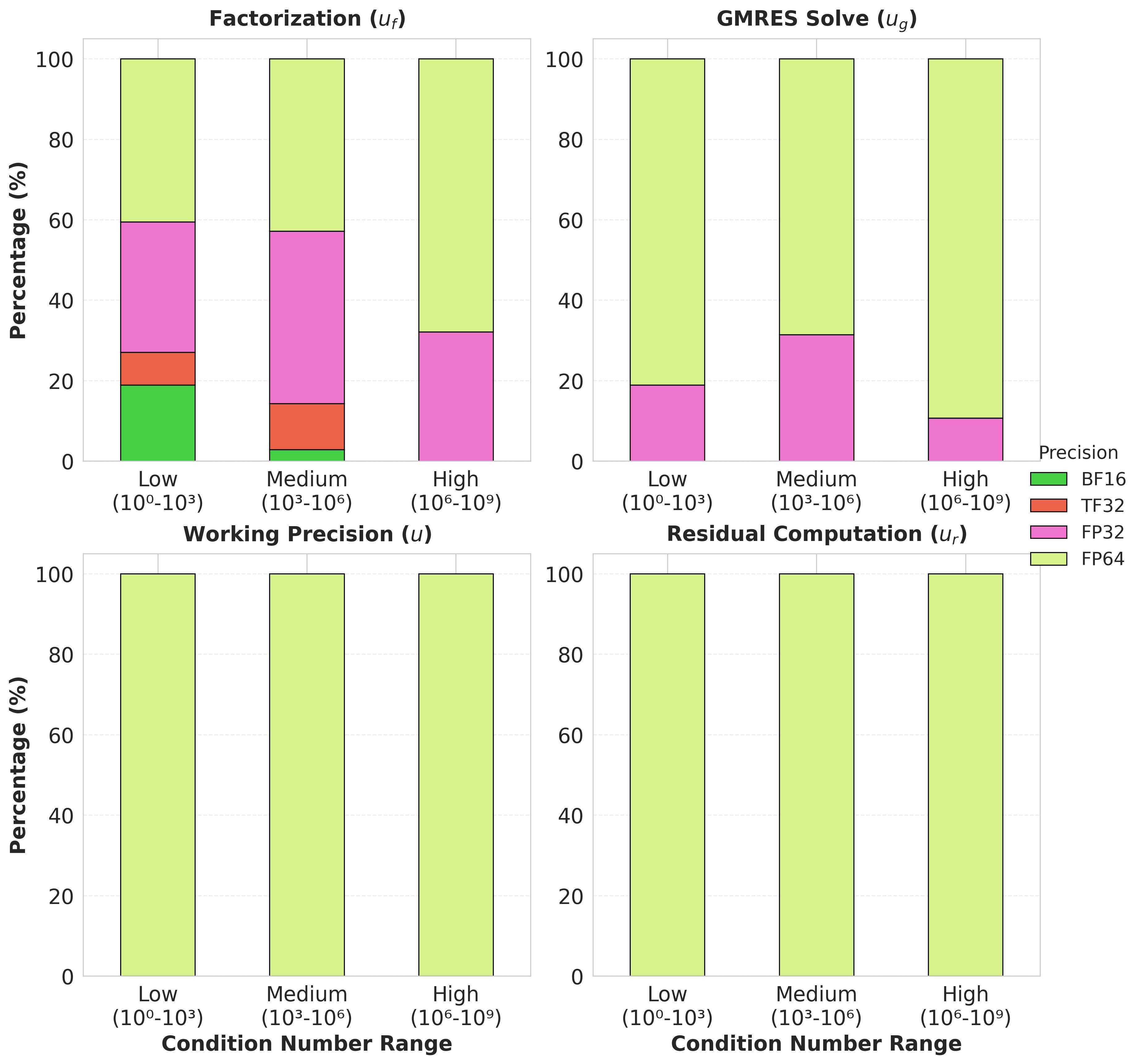}}
    \subfigure[RL ($W_2$, $\tau = 10^{-6}$)]{\includegraphics[width=0.43\linewidth]{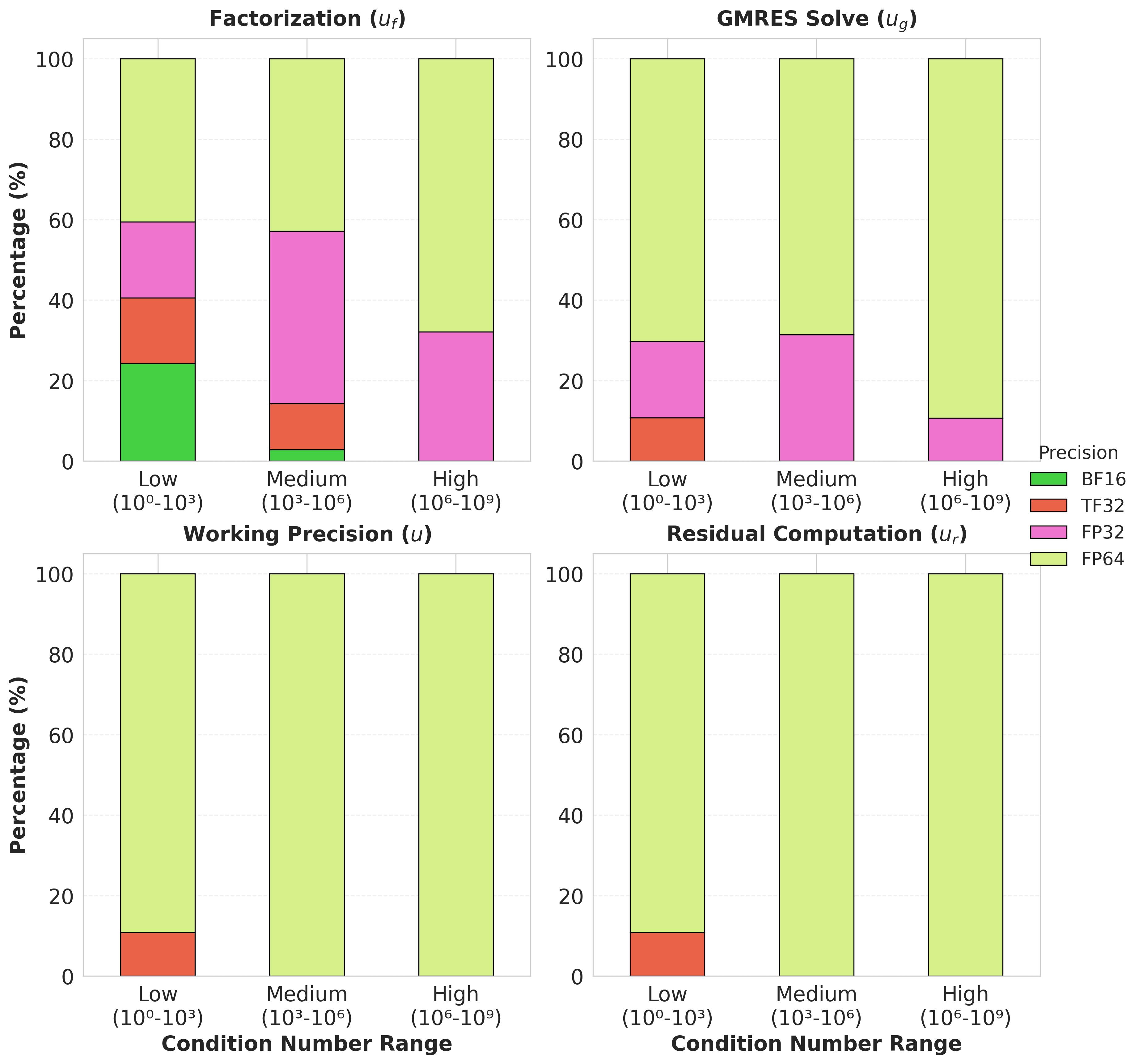}}
    \subfigure[RL ($W_1$, $\tau = 10^{-8}$)]{\includegraphics[width=0.43\linewidth]{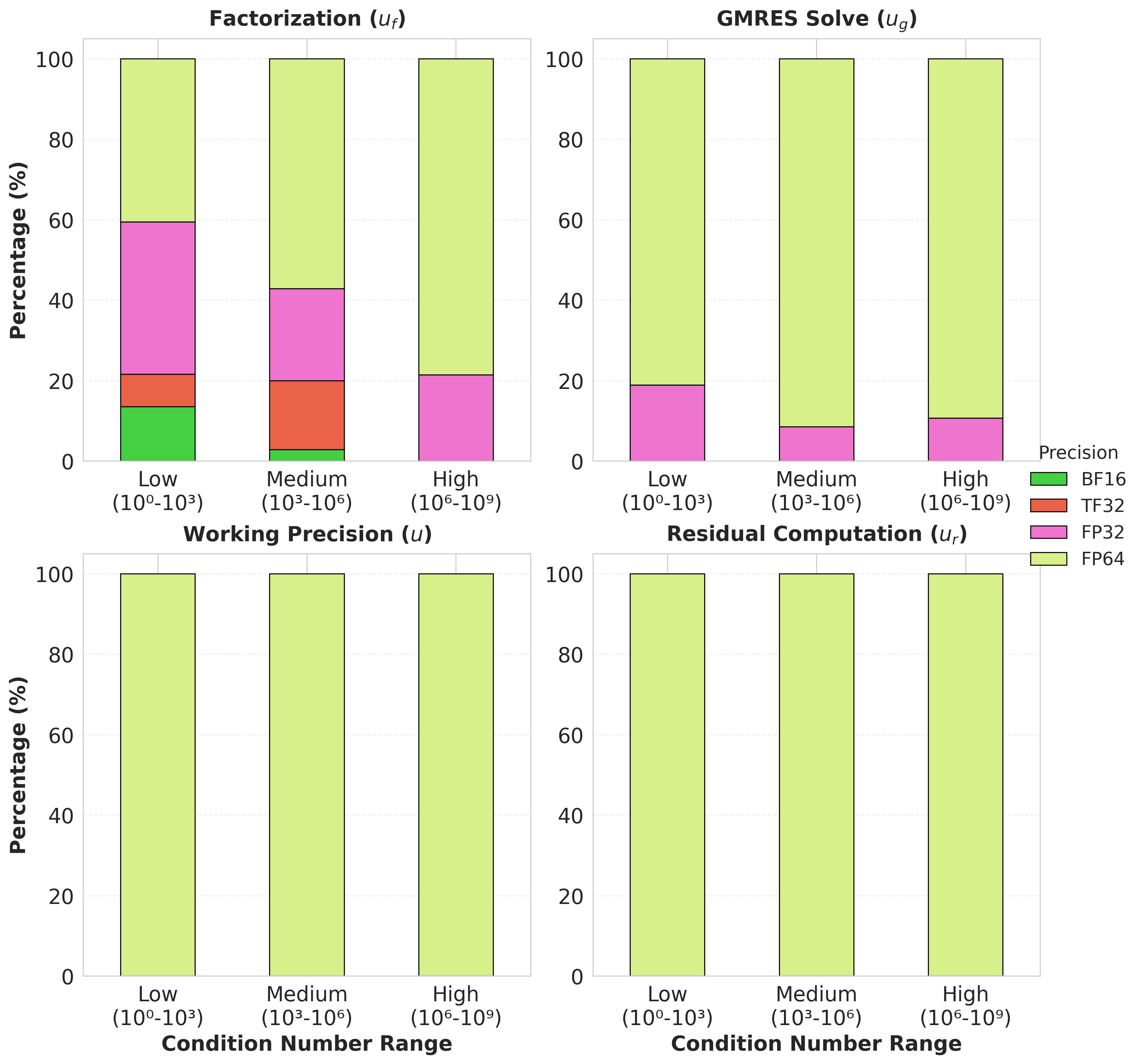}}
    \subfigure[RL ($W_2$, $\tau = 10^{-8}$)]{\includegraphics[width=0.43\linewidth]{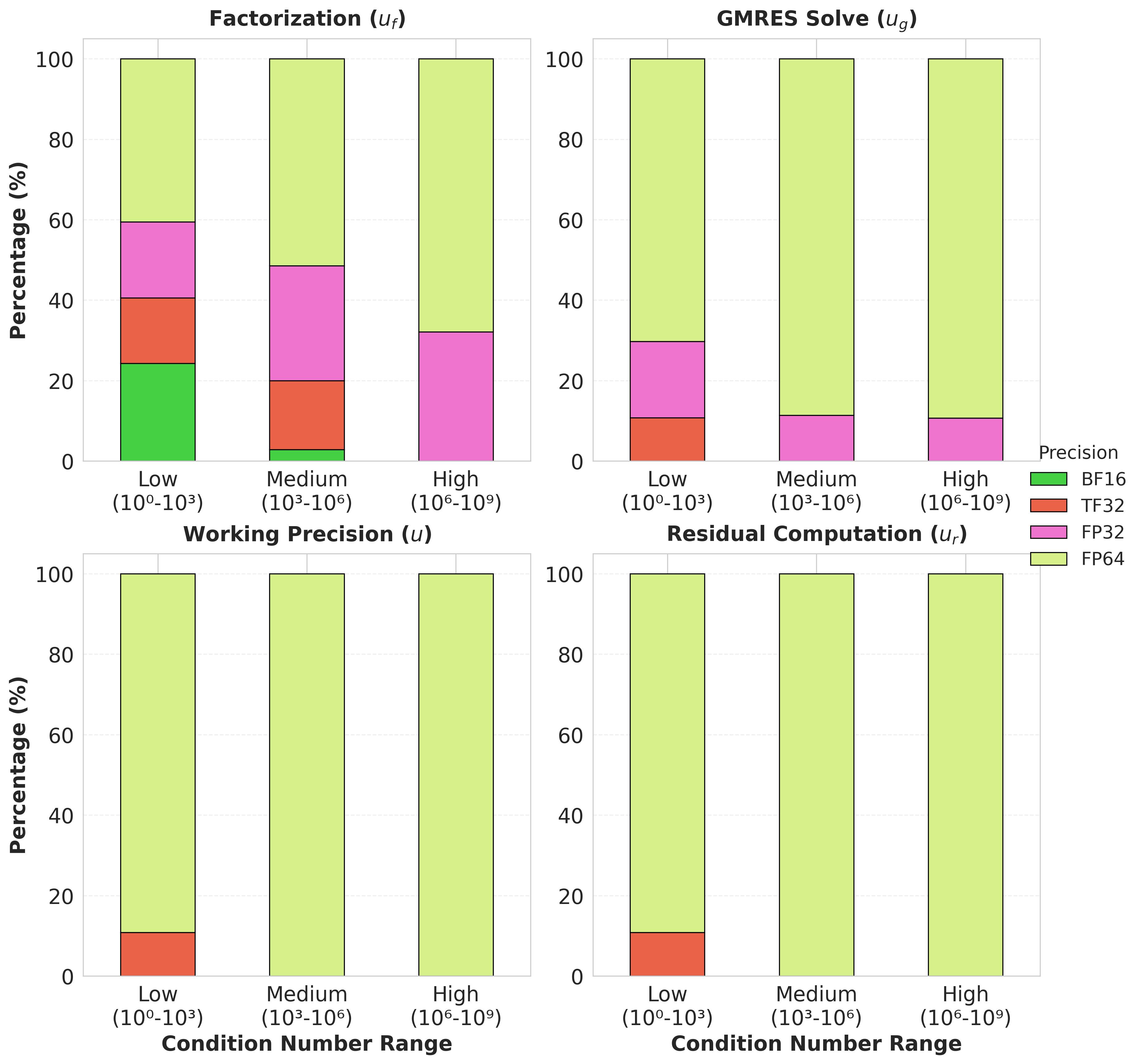}}
    \caption{Average floating-point formats selected across low, medium, and high condition-number ranges for the dense ablation study in which the iteration penalty $f_{\mathrm{penalty}}$ is removed from the reward. Each panel reports one reward-weight/tolerance setting: $W_1=(w_1,w_2)=(1,0.1)$ or $W_2=(1,1)$, with $\tau \in \{10^{-6},10^{-8}\}$.}
    \label{fig:pm_ws_ablation_usage}
\end{figure*}

\subsection{Native CPU validation}

\xy{A solver-level CPU validation is summarized in \tablename~\ref{tab:cpu_native_validation}. This validation uses a separate native C++ dense GMRES-IR implementation with the precision set $\{\text{FP16},\text{FP32},\text{FP64}\}$ for the four controlled stages $(u_f,u_g,u,u_r)$. The policy is trained on 50 dense systems drawn from the low- and medium-condition regimes and is validated on 100 dense systems spanning the low-, medium-, and high-condition regimes. Matrix sizes are sampled in the range 1000--1500. The low/medium/high labels are defined by the generator's controlled target condition number and are used only for dataset construction and reporting. The RL state does not observe this generator metadata; instead, it uses a Hager--Higham-type one-norm condition estimate computed from each generated matrix, together with $\|A\|_\infty$, so the selected action can vary across test matrices rather than being a single fixed precision tuple. For validation, every legal precision action is measured under the same fixed budget: three repeats, at most ten outer refinement steps, GMRES restart dimension 30, one restarted GMRES cycle per outer step, and inner relative tolerance $10^{-4}$. The reported row then uses only the action selected by the learned policy for each test matrix and compares it against the all-FP64 baseline.  All reported experiments and native validation runs were executed as CPU jobs on dual-socket AMD EPYC 7543 and Intel Xeon Gold 6330 processors. CPU timing is used for the native timing validation because it provides full control over all four GMRES-IR precision stages in a concise implementation. GPU timing is left outside the scope of this revision, since a fair GPU study would require a dedicated CUDA implementation of the entire stage-controlled GMRES-IR pipeline; otherwise measurements would be affected by library precision modes, data conversions, kernel launches, and Tensor Core eligibility.}

\begin{table*}[ht!]
\centering\scriptsize
\setlength{\tabcolsep}{4.2pt}
\renewcommand{\arraystretch}{1.15}
\caption{Native CPU GMRES-IR validation of the learned cross-regime policies. The ``Policy actions'' column reports the number of distinct four-stage precision tuples $(u_f,u_g,u,u_r)$ actually selected across the 100 validation matrices. The ``Reduced/mixed'' column reports how many validation cases use a selected action different from the all-FP64 tuple. Speedup and memory ratio are relative to the all-FP64 CPU baseline under the same stopping criteria.}
\label{tab:cpu_native_validation}
\begin{tabular}{l c c c c c c c c c}
\toprule
\textbf{Method} & {$\boldsymbol{\tau}$} & \textbf{Policy actions}
& \shortstack{\textbf{Reduced/}\\\textbf{mixed}}
& \shortstack{\textbf{Successful}\\\textbf{solves}}
& \textbf{Speedup}
& \textbf{Memory ratio}
& \shortstack{\textbf{Avg. ms}\\\textbf{(RL/FP64)}}
& \textbf{Avg. ferr}
& \textbf{Avg. nbe}\\
\midrule
RL ($W_1$) & $10^{-6}$ & 5 & 53/100 & 100/100 & $1.74\times$ & 0.92 & 227.73/320.09 & 1.84e-8 & 6.53e-9 \\
RL ($W_2$) & $10^{-6}$ & 5 & 53/100 & 100/100 & $1.74\times$ & 0.92 & 227.73/320.09 & 1.84e-8 & 6.53e-9 \\
RL ($W_1$) & $10^{-8}$ & 4 & 28/100 & 100/100 & $1.44\times$ & 0.91 & 268.97/320.15 & 5.68e-12 & 3.45e-12 \\
RL ($W_2$) & $10^{-8}$ & 5 & 29/100 & 99/100 & $1.45\times$ & 0.91 & 266.61/320.15 & 5.74e-12 & 3.49e-12 \\
\bottomrule
\end{tabular}
\end{table*}

\xy{The CPU validation shows that the learned policies transfer to a compiled solver implementation and produce time-to-solution improvements under native execution. In \tablename~\ref{tab:cpu_native_validation}, ``Successful solves'' is the CPU-validation counterpart of the success counts used above, but reported directly over the fixed 100-matrix validation set rather than as condition-binned rates. A CPU validation solve is counted as successful when the native run returns a successful solver status and the selected-policy relative forward error satisfies $\mathrm{ferr}\leq\tau$; the reported $\mathrm{nbe}$ values are additional backward-error diagnostics, not a separate acceptance rule for this timing table. Thus this column should not be read as an iteration count, but as the number of validation matrices satisfying the native CPU acceptance criterion. Three of the four learned policies solve all 100 validation systems, while the stricter $W_2,\tau=10^{-8}$ policy solves 99 out of 100 systems. The reported speedup is the arithmetic mean of
\begin{equation}\label{eq:per-matrix-speedup}
    \frac{t_{\mathrm{FP64}}(A_i)}{t_{\mathrm{RL}}(A_i)}
\end{equation}
over the successful validation matrices, where $t_{\mathrm{RL}}$ is the measured runtime of the action selected by the learned policy and $t_{\mathrm{FP64}}$ is the measured runtime of the all-FP64 baseline for the same matrix and stopping criteria. The ``Avg. ms'' column separately reports the arithmetic means of $t_{\mathrm{RL}}$ and $t_{\mathrm{FP64}}$ over the full validation set. The memory ratio is also averaged over the full validation set, but it is a model-estimated action-level storage ratio rather than a measured peak-memory value. The estimate counts the dense working matrix in precision $u$, an additional residual matrix when $u_r\neq u$, factor/preconditioner storage controlled by $u_f$, and vector workspaces controlled by $u_g$, $u$, and $u_r$, then normalizes by the corresponding all-FP64 estimate.

The ``Policy actions'' column is not an optimization target and should not be interpreted as diversity for its own sake; it records how many different precision tuples the state-dependent policy actually uses on the validation set. The ``Reduced/mixed'' column gives the complementary case-level view: it counts matrices for which the selected action is not $(\mathrm{FP64},\mathrm{FP64},\mathrm{FP64},\mathrm{FP64})$. The speedup column averages the per-matrix ratio~\eqref{eq:per-matrix-speedup} over the successful validation solves.
Therefore, matrices for which the policy selects the all-FP64 action remain in the average and contribute approximately unit speedup. Instead, the policies use reduced or mixed precision on $53\%$ of validation matrices at $\tau=10^{-6}$, and on $28\%$--$29\%$ at $\tau=10^{-8}$, while retaining the all-FP64 tuple as a fallback for the remaining cases. At $\tau=10^{-6}$, actions containing FP16 are selected on $20\%$ of the cases and actions containing FP32 are selected on $37\%$ of the cases for both $W_1$ and $W_2$. At $\tau=10^{-8}$, FP16 appears on $15\%$--$16\%$ of the cases and FP32 appears on $16\%$--$17\%$, reflecting a more conservative policy under the tighter tolerance. The selected action sets at $\tau=10^{-6}$ are
\[
\begin{aligned}
\mathcal{A}_{W_1,10^{-6}}=\mathcal{A}_{W_2,10^{-6}}=\{&
(\mathrm{FP64},\mathrm{FP64},\mathrm{FP64},\mathrm{FP64}),\\
&(\mathrm{FP32},\mathrm{FP32},\mathrm{FP32},\mathrm{FP64}),\\
&(\mathrm{FP16},\mathrm{FP64},\mathrm{FP64},\mathrm{FP64}),\\
&(\mathrm{FP16},\mathrm{FP32},\mathrm{FP64},\mathrm{FP64}),\\
&(\mathrm{FP16},\mathrm{FP16},\mathrm{FP32},\mathrm{FP32})\}.
\end{aligned}
\]
At the tighter tolerance, the selected sets become
\[
\begin{aligned}
\mathcal{A}_{W_1,10^{-8}}=\{&
(\mathrm{FP64},\mathrm{FP64},\mathrm{FP64},\mathrm{FP64}),\\
&(\mathrm{FP32},\mathrm{FP64},\mathrm{FP64},\mathrm{FP64}),\\
&(\mathrm{FP16},\mathrm{FP64},\mathrm{FP64},\mathrm{FP64}),\\
&(\mathrm{FP16},\mathrm{FP32},\mathrm{FP64},\mathrm{FP64})\},\\[2pt]
\mathcal{A}_{W_2,10^{-8}}=&\ \mathcal{A}_{W_1,10^{-8}}
\cup\{(\mathrm{FP16},\mathrm{FP16},\mathrm{FP32},\mathrm{FP32})\}.
\end{aligned}
\]
These sets show that the all-FP64 tuple remains available as a learned safe fallback, while the other selected tuples use FP16 or FP32 in selected stages when the discretized state indicates that reduced precision is admissible. The action set contracts as $\tau$ tightens from $10^{-6}$ to $10^{-8}$, and the reduced/mixed case fraction drops from $53\%$ to roughly $29\%$, which is consistent with the policy becoming more conservative under a stricter accuracy requirement.

Because the CPU policy is trained only on low- and medium-regime systems, the high-regime validation cases should be interpreted as an extrapolation test rather than as evidence that the policy has learned an optimized high-regime mixed-precision strategy. The agent does not observe the regime label; it observes the estimated condition feature. High-regime matrices therefore enter larger condition bins, where the learned policy is naturally conservative or falls back to the all-FP64 action when reduced-precision actions are not supported by the training updates. This behavior is desirable for validation: the policy does not blindly apply aggressive low precision outside the training distribution.

The mean successful-case speedups are $1.74\times$ for both $W_1$ and $W_2$ at $\tau=10^{-6}$, $1.44\times$ for $W_1$ at $\tau=10^{-8}$, and $1.45\times$ for $W_2$ at $\tau=10^{-8}$. The model-estimated storage ratio is approximately $0.91$--$0.92$ of FP64 across the four configurations. These results support the practical interpretation of the framework while remaining deliberately modest: the table reports measured solver runtime and an action-level storage model, not peak memory traffic or energy consumption.
}

\subsection{Study on Iteration Penalty}

The limitation of precision tuning in linear solvers without accounting for convergence iterations often appears in dynamic precision tuning tools such as HiFPTuner~\citep{10.1145/3213846.3213862}, because they focus primarily on precision tuning subject to accuracy constraints.

To further verify the effectiveness of our RL modeling for precision selection and performance balance,  we further conduct a reward component analysis to evaluate the contribution of the penalty term. We focus on the total converged GMRES iterations taken for iterative refinement and conduct an ablation study on training with \eqref{eq:reward} without the penalty $f_{\text{penalty}}$ in the iteration terms.

The results are presented in \tablename~\ref{tab:pm_ws_ablation}, together with \figurename~\ref{fig:pm_ws_ablation_usage} showing precision types chosen by RL. Similarly to above, we observed that between low- and medium-conditioned linear systems, RL strategically selects aggressive reduced-precision steps, whereas in high-conditioned linear systems it adopts a more conservative strategy for precision selection. Compared to the training with $f_{\text{penalty}}$ as shown in \figurename~\ref{fig:prec_usage}, removing $f_{\text{penalty}}$ leads to more reduced-precision actions. This removal increases the number of iterations required for the RL agent to achieve comparable accuracy, which means the iterative refinement uses extra iterations to compensate for the accuracy loss resulting from increased low-precision usage. The results indicate the importance of the reward function modeling for RL training and demonstrate the effectiveness of RL for precision selection under proper reward modeling.

\section{Discussion}
Unlike supervised learning problems, such as classification, our contextual bandit RL framework is well suited to practical numerical solvers because it does not require ground-truth labels for the optimal precision configuration. \xy{The reference solution used in the training reward is only used to evaluate the numerical error produced by a sampled precision action; it is not a label specifying which precision configuration should be chosen.} However, our RL framework for precision selection requires proper state modeling of the data. For instance, we use features such as the condition number and matrix norm in our state modeling for the linear solve problem. Selecting or modeling the states often requires expert knowledge to balance reasonable computational complexity with sufficient information for effective RL training.

However, this limitation can also be viewed as an advantage. In this work, the proposed framework has shown efficient precision selection even when prior knowledge about the required precision for critical computational steps is lacking. Although prior studies have explored mixed-precision usage in iterative refinement methods based on condition numbers and matrix sizes, our framework offers two unique benefits. First, it can quickly determine the required precision for unseen linear systems. Second, it allows us to identify the required precision for selected matrix features in a feature-based black-box setting with respect to precision labels. For example, by selecting certain features in our state space, we can examine whether these features are salient factors for determining reduced mixed precision. When the salient features are unknown or when their influence on the required precision level is unclear, our framework becomes particularly advantageous.

\section{Conclusion and Future Work}
\label{sec:conclusion}

We present a reinforcement learning framework for automated precision selection in linear solvers, casting the problem as a contextual bandit that balances convergence across iterations and low precision usage. Using iterative refinement as a case study, the framework dynamically adjusts precision at different computation steps to better balance numerical accuracy and efficiency. Our empirical results show that RL, leveraging a discretized state space, effectively adapts precision choices to the numerical properties of each linear system and achieves a good compromise between algorithm performance (relative error and converged iterations) and the reduced-precision portion of the linear systems under varying conditions.

In the future, we will explore deep RL for continuous-state spaces, incorporate additional significant matrix features (e.g., fill-in ratio, pivot growth estimates, sparsity, and spectral properties), and examine their contributions to improving the method's performance by reducing precision.  \xy{The native CPU validation evaluates learned FP16/FP32/FP64 configurations in a compiled GMRES-IR solver, so we report solver-level time-to-solution for this CPU implementation, but do not claim measured peak-memory traffic or energy reductions. A natural next step is to evaluate SuiteSparse and application matrices, include additional fixed mixed-precision GMRES-IR baselines, add a dedicated native implementation for architecture-specific timing studies, and replace the current bit-count proxy in the reward with platform-specific runtime, memory-traffic, or energy measurements.} By addressing these, we will be able to extend this framework to a broader range of numerical algorithms to advance the mixed-precision techniques.

\FloatBarrier

\begin{dci}
The authors declared no potential conflicts of interest with respect
to the research, authorship, and/or publication of this article.
\end{dci}

\begin{funding}
Erin Carson is supported by the Charles University Research Centre
program No. UNCE/24/SCI/005 and the European Union
(ERC, inEXASCALE, 101075632). Views and opinions expressed are
those of the authors only and do not necessarily reflect those of
the European Union or the European Research Council. Neither the
European Union nor the granting authority can be held responsible
for them.
\end{funding}

\begin{dataavailability}
The experimental code and data
used in this study are available at
\url{https://github.com/inEXASCALE/rlops}. 
\end{dataavailability}

\FloatBarrier

\bibliographystyle{SageH}
\bibliography{references}
\end{document}